\pdfoutput=1
\documentclass{article} %

\usepackage[sglblindworkshop,final]{neurips_2026}
\workshoptitle{FLLMPT}
\makeatletter\renewcommand{\@noticestring}{Accepted to the FLLMPT workshop, NeurIPS 2026.}\makeatother

\usepackage[utf8]{inputenc} %
\usepackage[T1]{fontenc}    %
\usepackage{amsmath,amsfonts,bm}

\def\eqref#1{equation~\ref{#1}}

\def\1{\bm{1}}

\DeclareMathAlphabet{\mathsfit}{\encodingdefault}{\sfdefault}{m}{sl}
\SetMathAlphabet{\mathsfit}{bold}{\encodingdefault}{\sfdefault}{bx}{n}

\usepackage[hidelinks]{hyperref}
\usepackage{url}
\usepackage{booktabs}
\usepackage{graphicx}
\usepackage{amsmath,amssymb}
\usepackage{nicefrac}
\usepackage{microtype}
\usepackage{xcolor}
\usepackage{placeins}

\newcommand{\kl}{\mathrm{KL}}
\newcommand{\dlogp}{\Delta\log p}

\title{A Privileged Teacher That Only Scores \\ Cannot Search: Critique-Conditioned \\ On-Policy Self-Distillation for Reasoning}

\author{%
  Mayug Maniparambil$^{1}$\thanks{Correspondence: \texttt{mayugmaniparambil@gmail.com}} \hspace{0.45em} Ayush Rai$^{2}$ \hspace{0.45em} Ekansh Verma$^{1}$ \hspace{0.45em} Noel O'Connor$^{2}$ \hspace{0.45em} Fergal Reid$^{1}$ \\
  $^{1}$Fin.ai \qquad $^{2}$Dublin City University
}

\begin{document}

\maketitle

\begin{abstract}

On-policy self-distillation promises to replace the sparse outcome signal of verifiable-reward RL with dense token-level credit: a teacher conditioned on the reference solution scores the student's own rollout, and the resulting distributional shift is distilled back into the policy. We investigate whether this scoring signal actually contains corrective credit. Using gold error spans in student rollouts, we test the two signals consumed by self-distillation objectives, the full-distribution KL and the token log-ratio, for two properties: whether they localize errors and whether they push probability away from erroneous student tokens. They exhibit neither of these properties nor can they reliably distinguish correct from incorrect rollouts within the same problem. We argue that this failure is structural: locating an error in a reasoning trace is itself a search problem, while a scoring teacher is never explicitly asked to perform that search. The same model, with the same privileged information, localizes errors substantially better when allowed to generate a critique. This forms the  motivation for \emph{critique-conditioned self-distillation}: first generate a critique of the current rollout, then condition the scoring teacher on that critique. The resulting signal localizes errors, is corrective within error spans, and shifts probability toward tokens that initiate backtracking such as ``Wait'', ``But'', and ``Actually''. We further show that the critique is useful both at test time, where resuming a wrong rollout with the critique in context converts math problems the model cannot solve by sampling, and in cold-start GRPO, where it provides an advantage signal on prompts for which verifiable rewards alone provide none.

\end{abstract}

\section{Introduction}

Reinforcement learning with verifiable rewards (RLVR) provides a reliable but sparse learning signal: a reasoning rollout is correct or incorrect, but the verifier does not identify which intermediate decisions caused the outcome. In group-based objectives, the limitation is sharper when all sampled rollouts for a problem fail, since the relative advantage can vanish entirely. These all-wrong groups are precisely where additional supervision is most useful, yet outcome-based RL provides little guidance about how the reasoning should change.

On-policy self-distillation (OPSD) offers a natural way to obtain denser supervision
\citep{zhao2026selfdistilledreasoneronpolicyselfdistillation,hubotter2026reinforcementlearningselfdistillation,yang2026selfdistilledrlvr}.
The model generates a rollout as a student and then scores the same rollout as a teacher with privileged information, such as a reference solution. The discrepancy between the student and teacher distributions provides a signal at every token, even when the final answer is wrong. But density alone does not imply credit assignment. For the signal to be corrective, it should be larger where the reasoning goes wrong and should push probability away from the student's erroneous choices.

We test these properties directly on annotated error spans in logic and competition-mathematics rollouts. Across tasks and teacher scales, reference-conditioned scoring fails both tests: its distributional shift is only weakly concentrated at errors, its token-level signal does not reliably suppress erroneous student tokens, and even when aggregated over the full trajectory it does not distinguish correct from incorrect rollouts of the same problem. The limitation is therefore not in how a particular loss consumes the signal. Scoring itself exposes little information about where the reasoning failed.

We argue that locating an error is itself a reasoning problem. A scoring teacher is given the reference but is never explicitly asked to compare it against the rollout and search for the mistake. When the same model is instead allowed to generate a critique, it can localize and explain the error. This motivates \emph{critique-conditioned self-distillation}: first generate a localized critique of the student's rollout, then condition the scoring teacher on that critique. Generation performs the search, and scoring converts the error it finds into a dense token-level signal.

Our contributions are threefold. We show that the privileged teacher's signal is neither localized nor corrective at gold error spans, and is not predictive of rollout correctness even at the whole-trace level (\S\ref{sec:empty}). We show that the failure lies in scoring, not in the information. The same model, given the same reference, localizes errors when it generates a critique (\S\ref{sec:search}). Finally, we introduce \emph{critique-conditioned self-distillation}, which uses generated critiques as privileged context for dense token-level supervision, and provide initial results using this signal for test-time inference and training (\S\ref{sec:using}, Appendix~\ref{app:training}).

\section{Related Work}
\label{sec:related}

\textbf{Privileged self-distillation.} Our work sits between privileged self-distillation and error-localization methods. On-policy self-distillation uses a teacher conditioned on privileged information to provide dense token-level supervision over the student's own rollout \citep{zhao2026selfdistilledreasoneronpolicyselfdistillation,hubotter2026reinforcementlearningselfdistillation,yang2026selfdistilledrlvr,pan2026rlcsdreinforcementlearningcontrastive}. Recent work has identified several failure modes of this setup, including context-induced bias and weak or stylistic distributional shifts \citep{harne2026privilegedbiasedpiconditionedteachers,kaur2026rethinkingonpolicyselfdistillationthinking,pan2026rlcsdreinforcementlearningcontrastive}. We differ by measuring the resulting signal directly at gold error spans and asking whether it carries corrective credit.

\textbf{Locating errors is a reasoning problem.} A separate line of work shows that locating reasoning errors can itself require generation or explicit reasoning, motivating generative verifiers, process supervision, and critique-based reward models \citep{tyen2024llmsreasoningerrorscorrect,zhang2025generativeverifiersrewardmodeling,ankner2024critiqueoutloudrewardmodels,khalifa2025processrewardmodelsthink}. Extended related work appears in Appendix~\ref{app:related}.

\section{Preliminaries}
\label{sec:prelim}

\subsection{Group-relative RL and the all-wrong group}
\label{sec:prelim-grpo}
For a prompt $x$ the policy $\pi_\theta$ samples a group of $G$ rollouts
$y^1,\dots,y^G$, a verifier scores each one $r_i\in\{0,1\}$, and every token of
rollout $i$ receives the group-normalised advantage
$A_i=(r_i-\bar r)/\sigma_r$ \citep{shao2024deepseekmathpushinglimitsmathematical}. The signal is one
bit per rollout, and it vanishes entirely when the group is unanimous: if all
$G$ rollouts are wrong, $A_i=0$ for every $i$ and the group contributes no
gradient. These \emph{all-wrong} groups are the problems at the frontier of the
model's competence---the ones it cannot yet solve once in $G$ tries---and they
are the reason verifiable-reward RL tends to sharpen what the base model can
already do rather than extend it \citep{yue2025doesreinforcementlearningreally}.

\subsection{On-policy self-distillation}
\label{sec:prelim-opsd}

Let $P_S(\cdot\mid x,y_{<t})$ denote the student's next-token distribution at a prefix of its own rollout, and let $P_T(\cdot\mid x,r,y_{<t})$ denote the same network evaluated at that prefix with privileged information $r$---here, the reference solution---in context. Because the teacher evaluates a rollout it did not generate, we call this \emph{scoring}. Scoring yields a dense signal: at each position, the discrepancy between $P_T$ and $P_S$ is interpreted as token-level credit.

Two forms of that discrepancy are in use. OPSD distils the full distribution,
\begin{equation}
\mathcal L_{\mathrm{KL}}=\sum_t \kl\!\big(P_T(\cdot\mid x,r,y_{<t})\,\big\|\,P_S(\cdot\mid x,y_{<t})\big),
\label{eq:opsd-kl}
\end{equation}
so its per-token signal is the KL divergence at each position
\citep{zhao2026selfdistilledreasoneronpolicyselfdistillation,hubotter2026reinforcementlearningselfdistillation,kaur2026rethinkingonpolicyselfdistillationthinking}.
RLSD \citep{yang2026selfdistilledrlvr} instead scores the token the student wrote,
\begin{equation}
\Delta_t=\log P_T(y_t\mid x,r,y_{<t})-\log P_S(y_t\mid x,y_{<t}),
\label{eq:opsd-lr}
\end{equation}
and uses $\Delta_t$ to modulate the group advantage token by token. The verifier therefore determines the update direction, while the privileged teacher modulates its magnitude.
RLCSD \citep{pan2026rlcsdreinforcementlearningcontrastive} keeps this form but replaces the reference by sibling rollouts from the same group, contrasting a correct sibling $y^{+}$ against a wrong one $y^{-}$, $e_t=\log P_T(y_t\mid x,y^{+},y_{<t})-\log P_T(y_t\mid x,y^{-},y_{<t})$, thereby removing shifts induced by privileged context alone. Neither construction resolves the all-wrong regime: there is no correct sibling, and group-relative advantage vanishes. We therefore focus on reference-conditioned scoring and ask a more basic question: does $P_T(\cdot\mid x,r,y_{<t})$ expose useful credit at the positions where the student's reasoning goes wrong?

\subsection{A general form of the probe}
\label{sec:prelim-probe}

Fix a rollout $y$ and let $b$ denote privileged context prepended to the teacher prompt. Scoring the same rollout with and without $b$ gives two per-token signals:
\begin{align}
\mathrm{KL}_t(b)
&=
\kl\!\big(
P_T(\cdot \mid x,b,y_{<t})
\,\big\|\,
P_T(\cdot \mid x,y_{<t})
\big),
\label{eq:probe-kl}\\
\Delta_t(b)
&=
\log P_T(y_t \mid x,b,y_{<t})
-
\log P_T(y_t \mid x,y_{<t}).
\label{eq:probe-lr}
\end{align}
When $b=r$, these recover the two OPSD readouts of Eqs.~\ref{eq:opsd-kl} and \ref{eq:opsd-lr}, since the unprivileged teacher is the student.

The two signals probe different properties of the same scoring pass. $\mathrm{KL}_t(b)$ measures how strongly the privileged context changes the teacher's distribution at position $t$. If that context identifies where the rollout fails, this shift should concentrate at the error spans, so we call it the \emph{localization} signal. $\Delta_t(b)$ measures how the context changes the probability of the token the student actually wrote. At an erroneous token, a corrective signal should reduce that probability, so we call $\Delta_t$ the \emph{correction} signal and read negative values as corrective. We evaluate both signals at gold error spans in \S\ref{sec:setup}.

To isolate the information carried by privileged context, we pair each context $b$ with a matched placebo $\tilde b$ and report the content signals $\mathrm{KL}_t(b)-\mathrm{KL}_t(\tilde b)$ and $\Delta_t(b)-\Delta_t(\tilde b)$. This controls for distributional shifts caused by the presence of a solution-shaped preamble alone. For the reference condition, $b=r$ and $\tilde b=r'$ is a matched-length reference from another problem. For the critique condition, $b=c(y)$ and $\tilde b=c(y')$ is a critique of another problem's rollout.

The critique is generated from the problem, reference, and student rollout as $c(y)=\{(q_j,\rho_j)\}_j$, where $q_j$ is a verbatim quote of an erroneous sentence and $\rho_j$ explains the error. \emph{Critique-conditioned self-distillation} then replaces the reference in OPSD with $c(y)$: generation performs the search for the error, and scoring converts the resulting critique into a dense signal over the rollout. We ablate the quote and explanation separately in \S\ref{sec:pointing}.

\section{Setup}
\label{sec:setup}

\subsection{Models, data, and labels}
\label{sec:setup-data}

\paragraph{Models and tasks.}
We conduct the token credit signal analysis on the Qwen3-1.7B
\citep{yang2025qwen3technicalreport}, Olmo-3-7B-Think and Olmo-3-32B-Think
\citep{olmo2026olmo3} rollouts. In each case the teacher is the same model with
privileged context, the on-policy self-distillation setting. For Qwen3-1.7B we
also run Qwen3-8B as a scale-control teacher. The tasks are Knights \& Knaves
logic puzzles \citep{xie2025memorizationlargelanguagemodels,xie2025logicrlunleashingllmreasoning}
(K\&K, 4--8 people) and DeepMath
\citep{he2025deepmath103klargescalechallengingdecontaminated} at two difficulty
bands (L5-7, L7-10).

\paragraph{Rollouts, references, and gold error spans.}
We study wrong rollouts: verifier-incorrect, not truncated, and ending in a
boxed answer. We keep the verifier-correct rollouts of the same problems for
the correctness test in \S\ref{sec:predictive}. The privileged reference is a
verified step-by-step derivation for K\&K and the dataset's reference solution
plus answer for DeepMath. We obtain gold error spans from a grader model that
reads the problem, the rollout and the reference and flags each error as a
verbatim quote of one sentence plus a reason. The grader is a DeepSeek model
\citep{deepseekai2026deepseekr1incentivizingreasoningcapability,deepseekai2026deepseekv4highlyefficientmilliontoken}
for the Qwen rollouts and claude-sonnet-5 \citep{anthropic2026claudesonnet5} for the Olmo-3 rollouts. We then have
gpt-5.2 \citep{openai2025gpt52} and claude-sonnet-5 judge every flag independently, and we count a
flag as gold when both accept it. To measure how lenient the judges are, we
replace a quarter of the flags sent to each judge with position-matched
control sentences carrying the same reason and report their acceptance rate
with each study (Appendix~\ref{app:setup}). We also perform the same signal
analysis on human-annotated error spans from ProcessBench (Appendix~\ref{app:human}).

\paragraph{Critics.}
We use three critics to write the critiques $c(y)$ of \S\ref{sec:prelim-probe}.
All use the same prompt: the rollout is presented sentence-numbered, each error
is a verbatim quote plus a reason, and recall is favoured over precision
(Appendix~\ref{app:prompt-critique}). All receive the same reference $r$ that
the teacher receives in the reference condition. We use gpt-5.2 as the oracle critic.
We also test whether the student can identify its own errors by having Qwen3-1.7B critique its own rollouts. We call this the self-critique. To assess whether a larger critic is more effective, we have Qwen3-8B perform the critiques as well. Appendix~\ref{app:examples} shows an oracle critique and a self-critique of the same rollout on each task.
Because critic and teacher see the same information, the comparison in
\S\ref{sec:search} is between generating and scoring, not between amounts of
information. The Olmo-3 models each critique their own rollouts
(Table~\ref{tab:spanz}, Appendix~\ref{app:olmo}).

\subsection{The error-span Z readout}
\label{sec:setup-spanz}

Per-token signals vary substantially across positions and rollouts, so we normalise each rollout against its own clean regions, following baseline-normalisation conventions in event-related neurophysiology \citep{mercier2022advances}. For each rollout, we measure the signal in a short window immediately after the first gold error span and compare it with length-matched windows after clean spans in the prefix. We use the first error only because the reasoning after it is often wrong in ways the gold spans do not mark. If $\bar{s}_W$ denotes the mean signal in window $W$, we define
\[
Z_r
=
\frac{
\bar{s}_{W_{\mathrm{err}}}
-
\operatorname{mean}_k \bar{s}_{W_k}
}{
\operatorname{sd}_k \bar{s}_{W_k}
},
\]
which we call the \emph{error-span Z}. It measures how far the signal at the error departs from the rollout's own baseline, in units comparable across signals, teachers, and tasks.

A positive localization score means that the teacher's distribution changes more at the error than at clean regions while a negative correction score means that privileged context lowers the probability of the student's erroneous token. For localization, we log-transform the window-mean KL before computing $Z_r$ to reduce the effect of its heavy tail. We report the mean $Z_r$ across rollouts, calibrated by subtracting the score obtained when a random clean sentence from the same rollout is treated as the error. Thus, a signal with no error-specific content is centered at zero. Window sizes and other implementation details are given in Appendix~\ref{app:setup}.

\section{The Privileged Signal Is Empty Where It Matters}
\label{sec:empty}

\begin{figure}[t]
\centering
\includegraphics[width=0.75\linewidth]{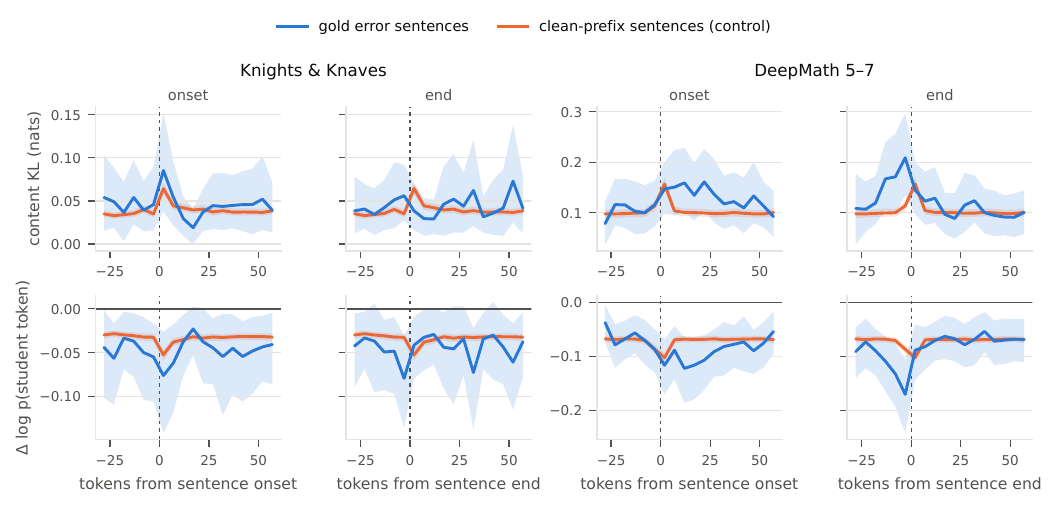}
\caption{\textbf{The reference-conditioned teacher barely reacts at the student's errors.} Content KL (top) and $\dlogp$ of the student's token (bottom) around gold error sentences (blue) and clean sentences (orange), aligned to sentence onset and end; Qwen3-1.7B scoring its own rollouts. Bands: 95\% CI.}
\label{fig:localization}
\end{figure}

We report the localization and correction signals of \S\ref{sec:prelim-probe}
at the positions of the student's rollouts known to be wrong.
Figure~\ref{fig:localization} shows both signals around the gold error
sentences (blue), aligned to sentence onset and to sentence end and averaged
over all error sentences of all rollouts, against the same average over the
clean sentences that precede the first error (orange). Table~\ref{tab:spanz}
reports the error-span Z of \S\ref{sec:setup-spanz}. A privileged teacher
provides usable token-level credit at an error if its error-span Z is positive
for localization and negative for correction. We apply the same criterion to
the reference here and to the critique in \S\ref{sec:deliver}.

\paragraph{Not localized at the errors.}
\label{sec:notlocalized}

On K\&K, error and control traces are nearly indistinguishable for both localization and correction. On DeepMath, the reference induces a small error-specific shift, most visible near the sentence boundary, but the effect is much weaker than under critique conditioning (Figure~\ref{fig:critique}). Thus, the reference does affect the teacher's distribution, but not selectively at the student's errors.

Table~\ref{tab:spanz} shows the same comparison at the rollout level. Across students, tasks, and both signals, reference-conditioned scores remain close to the rollout's clean baseline, and no setting satisfies the corrective criterion. When the reference signal does depart from zero, it can even point in the wrong direction: on L7-10, the Qwen teacher assigns \emph{more} probability to the student's erroneous token when conditioned on the reference. Increasing teacher capacity does not rescue the signal. Re-scoring the same Qwen rollouts with the 8B teacher leaves the reference-conditioned scores similarly small, while critique-conditioned signals become stronger across settings (Appendix~\ref{app:spanz}).

\begin{table}[t]
\centering
\caption{\textbf{Error-span Z at the first gold error, for three students each scoring its own rollouts.} Localization: log content KL (positive = reacts at the error); correction: $\dlogp$ of the student's token (negative = corrective). Bold: 95\% CI excludes 0.}
\label{tab:spanz}
\small
\setlength{\tabcolsep}{4.5pt}
\begin{tabular}{lrrrrrr}
\toprule
 & \multicolumn{2}{c}{K\&K} & \multicolumn{2}{c}{L5-7} & \multicolumn{2}{c}{L7-10} \\
\cmidrule(lr){2-3}\cmidrule(lr){4-5}\cmidrule(lr){6-7}
conditioning & loc & corr & loc & corr & loc & corr \\
\midrule
\multicolumn{7}{l}{\emph{Qwen3-1.7B} (355 / 516 / 794 problems)} \\
reference & $\mathbf{-0.17}$ & $+0.11$ & $+0.14$ & $-0.14$ & $+0.07$ & $\mathbf{+0.17}$ \\
critique, oracle & $\mathbf{+0.75}$ & $\mathbf{-0.62}$ & $\mathbf{+0.99}$ & $\mathbf{-0.90}$ & $\mathbf{+0.78}$ & $\mathbf{-0.54}$ \\
critique, self & $+0.16$ & $-0.07$ & $\mathbf{+0.25}$ & $-0.14$ & $\mathbf{+0.18}$ & $-0.07$ \\
\midrule
\multicolumn{7}{l}{\emph{Olmo-3-7B-Think} (197 / 398 / 631 problems)} \\
reference & $\mathbf{+0.20}$ & $-0.00$ & $+0.11$ & $-0.08$ & $-0.01$ & $-0.03$ \\
critique, oracle & $\mathbf{+0.63}$ & $\mathbf{-0.29}$ & $\mathbf{+0.89}$ & $\mathbf{-0.59}$ & $\mathbf{+0.88}$ & $\mathbf{-0.49}$ \\
critique, self & $+0.13$ & $-0.14$ & $\mathbf{+0.41}$ & $\mathbf{-0.18}$ & $\mathbf{+0.33}$ & $\mathbf{-0.14}$ \\
\midrule
\multicolumn{7}{l}{\emph{Olmo-3-32B-Think} (114 / 111 / 133 problems)} \\
reference & $+0.09$ & $\mathbf{-0.41}$ & $+0.21$ & $-0.09$ & $+0.15$ & $-0.06$ \\
critique, oracle & $\mathbf{+0.76}$ & $\mathbf{-0.43}$ & $\mathbf{+0.96}$ & $\mathbf{-0.50}$ & $\mathbf{+1.20}$ & $\mathbf{-0.62}$ \\
critique, self & $\mathbf{+0.47}$ & $-0.01$ & $\mathbf{+0.54}$ & $-0.39$ & $\mathbf{+0.40}$ & $\mathbf{-0.42}$ \\
\bottomrule
\end{tabular}
\end{table}

\paragraph{Not even informative about correctness.}
\label{sec:predictive}

\begin{figure}[t]
\centering
\includegraphics[width=0.75\linewidth]{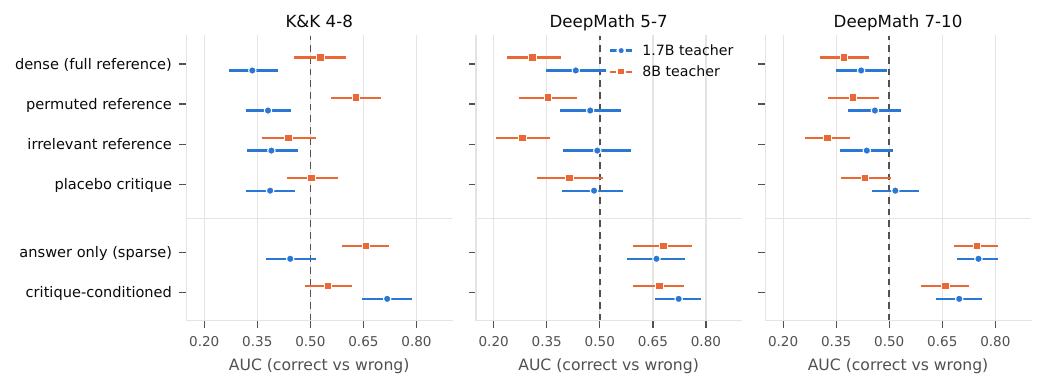}
\caption{\textbf{The reference-conditioned signal cannot tell correct rollouts from wrong ones.} Pooled AUC of trace-mean $\dlogp$, correct vs.\ wrong rollouts, by privileged context and teacher.}
\label{fig:predictive}
\end{figure}

A diffuse signal could still be useful even if it does not localize errors, but at minimum it should recover the verifier's one bit: whether the rollout is correct. We therefore test whether the localization or correction signal averaged over all tokens of a rollout ranks correct rollouts above incorrect siblings of the same problem. Across all three tasks and both teacher scales, neither reference-conditioned signal does so reliably: all twelve within-problem AUC confidence intervals include chance. Pooled across problems, some cells move away from chance (Figure~\ref{fig:predictive}), but the same deviations appear under content-free controls, indicating that they largely reflect problem-level factors such as length and difficulty. A trained probe over richer trace-level summaries does no better than the content-free controls (Appendix~\ref{app:pred}). The reference-conditioned signal is therefore not only poorly localized. Even aggregated over the full trajectory, it carries little information about whether the reasoning is correct.

\section{Dense Feedback Is a Search Problem}
\label{sec:search}

Why does the reference as privileged context induce so little signal at the error? We posit that locating where a rollout goes wrong, and why, is itself a reasoning problem: the reference and rollout must be compared until the discrepancy is found. A scoring pass is never asked to perform that search. It only produces a next-token distribution with the reference in context. If the failure is one of scoring rather than of information, the same model should recover the errors when allowed to generate. We test this directly by asking critics of different scales to read the problem, reference, and rollout and produce a list of errors, each with a verbatim quote and an explanation (\S\ref{sec:prelim-probe}). The quote makes the prediction directly localizable against the gold error spans and later provides an anchor for critique-conditioned teacher scoring.

Table~\ref{tab:gate} confirms this prediction. The strongest critic localizes a gold error on well over half of the rollouts across tasks, while smaller critics succeed less often but preserve the same ordering with scale. Their main failures are practical rather than conceptual: they often paraphrase instead of quoting the rollout or violate the required output format. The key result is that the same reference that yields little error-specific signal under scoring produces explicit error locations under generation. This suggests that search is the missing computation. The next section asks whether these generated critiques can in turn be converted back into a dense signal by scoring.

\begin{table}[t]
\centering
\caption{\textbf{Generating finds the errors that scoring cannot.} Percent of wrong rollouts with a critique quote within five sentences of a gold error (localized $\pm5$), quotes found verbatim in the rollout (quote-match), and critiques in the required format (parse).}
\label{tab:gate}
\footnotesize
\setlength{\tabcolsep}{3.5pt}
\begin{tabular}{lccc ccc ccc}
\toprule
 & \multicolumn{3}{c}{localized $\pm5$} & \multicolumn{3}{c}{quote-match} & \multicolumn{3}{c}{parse} \\
\cmidrule(lr){2-4}\cmidrule(lr){5-7}\cmidrule(lr){8-10}
critic & K\&K & L5-7 & L7-10 & K\&K & L5-7 & L7-10 & K\&K & L5-7 & L7-10 \\
\midrule
gpt-5.2 (oracle) & 57 & 62 & 67 & 98 & 96 & 95 & 100 & 100 & 100 \\
Qwen3-8B & 26 & 22 & 27 & 96 & 82 & 83 & 99 & 86 & 85 \\
Qwen3-1.7B (self) & 20 & 18 & 21 & 77 & 64 & 68 & 90 & 83 & 83 \\
\bottomrule
\end{tabular}
\end{table}

\section{Critique-Conditioned Self-Distillation Delivers}
\label{sec:deliver}

\begin{figure}[t]
\centering
\includegraphics[width=0.75\linewidth]{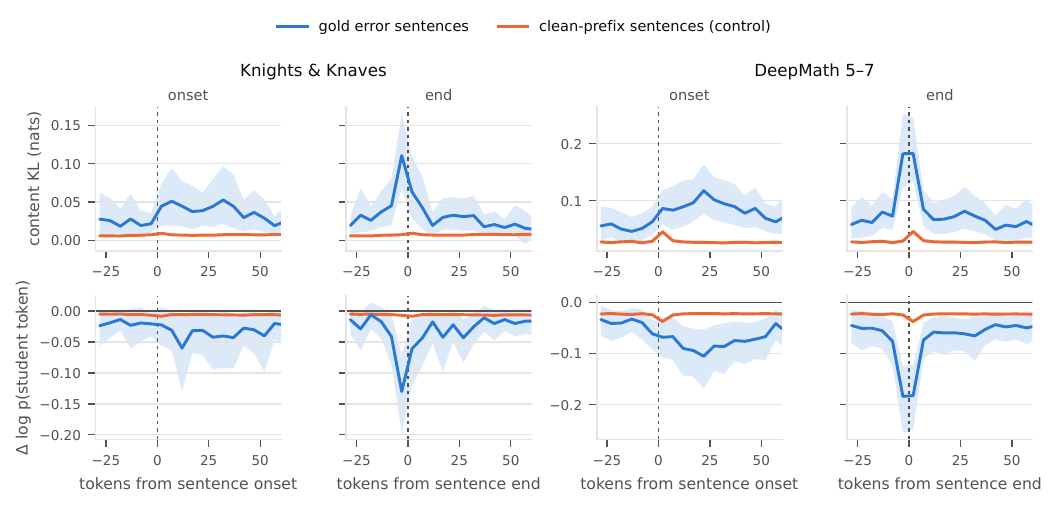}
\caption{\textbf{With a critique in context, the same teacher reacts at the errors.} Same layout as Figure~\ref{fig:localization}, with gpt-5.2 critiques in place of the reference; Qwen3-1.7B scoring its own rollouts. The error curve leaves its control on every panel.}
\label{fig:critique}
\end{figure}

We replace the reference in the teacher's context with a generated critique and score the student's rollout exactly as before, using a critique from another problem as the placebo. Figure~\ref{fig:critique} repeats Figure~\ref{fig:localization} with the critique in place of the reference. The teacher, alignment, and estimand are the same. Under critique conditioning, the previously overlapping error and control traces separate. The teacher's distribution shifts after the flagged error sentence, and its probability on the student's continuation decreases there rather than elsewhere. This pattern holds across tasks.

Table~\ref{tab:spanz} confirms the same result at the rollout level. With oracle critiques, every setting satisfies the corrective criterion of \S\ref{sec:empty} on both signals: errors are localized relative to clean regions, and probability is pulled away from the student's erroneous continuation, at several times the magnitude of the reference-conditioned signal. The critic sees only the reference that the scoring teacher already has. The critique therefore carries the same information in a form the scoring pass can use. Self-generated critiques preserve the same effect at reduced magnitude: with Qwen3-1.7B critiquing its own rollouts, localization is positive and correction is negative on all three tasks, and the effect strengthens with model scale. The weaker self-critique signal still trains. In the training runs of Appendix~\ref{app:training}, the self-critique arm recovers most of the oracle-critique arm's gain over GRPO (Table~\ref{tab:kkshort}). Larger scoring teachers likewise amplify the critique-conditioned signal while leaving the reference-conditioned signal near baseline (Appendix~\ref{app:spanz}). Critique conditioning also recovers the trace-level correctness signal absent under the reference (Figure~\ref{fig:predictive}). The same pattern holds on human-labelled error steps: on ProcessBench \citep{zheng2025processbenchidentifyingprocesserrors}, with each solution scored by the Qwen2.5 model that wrote it, the reference is neither localized nor corrective, while a critique localizes the labelled step for both generators and corrects once the tokens it quotes are excluded (Appendix~\ref{app:human}).

\paragraph{Correction resolves into self-correction language.}
\label{sec:reconsider}

The correction signal in Table~\ref{tab:spanz} shows that critique conditioning moves probability mass away from the student's continuation, but not where that mass goes. We therefore inspect high-content tokens in the post-error window and bucket the teacher's argmax prediction by destination. Under reference conditioning, error windows show no excess of self-correction markers such as \emph{Wait}, \emph{But}, or \emph{However} while under critique conditioning, the reconsideration bucket rises to two to three times the control rate on every task (Table~\ref{tab:reconsider}). Thus, the probability mass released from the student's erroneous continuation is redirected toward self-correction language only when the teacher is told what is wrong, not merely shown the correct solution.

\begin{table}[t]
\centering
\begin{minipage}[t]{0.37\linewidth}
\centering
\caption{\textbf{Critique conditioning redirects probability toward self-correction markers; the reference does not.} Share of post-error tokens whose argmax becomes a self-correction marker, at errors (err) and controls (ctl).}
\label{tab:reconsider}
\small
\begin{tabular}{lrrrr}
\toprule
 & \multicolumn{2}{c}{reference} & \multicolumn{2}{c}{critique} \\
\cmidrule(lr){2-3}\cmidrule(lr){4-5}
task & err & ctl & err & ctl \\
\midrule
K\&K  & 4.9 & 4.9 & 23.9 & 7.7 \\
L5-7  & 3.7 & 2.9 & 9.0  & 5.2 \\
L7-10 & 2.6 & 2.6 & 10.6 & 4.8 \\
\bottomrule
\end{tabular}
\end{minipage}\hfill
\begin{minipage}[t]{0.60\linewidth}\setlength{\tabcolsep}{3pt}
\centering
\caption{\textbf{The quote carries localization, the reason carries correction.} Error-span Z with the oracle's critiques ablated to quote only or reason only; Qwen3-1.7B teacher unless stated.}
\label{tab:ablation}
\resizebox{\linewidth}{!}{%
\begin{tabular}{lrrrrrr}
\toprule
 & \multicolumn{2}{c}{K\&K} & \multicolumn{2}{c}{L5-7} & \multicolumn{2}{c}{L7-10} \\
\cmidrule(lr){2-3}\cmidrule(lr){4-5}\cmidrule(lr){6-7}
critique & loc & corr & loc & corr & loc & corr \\
\midrule
quote + reason & $+0.75$ & $-0.62$ & $+0.99$ & $-0.90$ & $+0.78$ & $-0.54$ \\
reason only & $+0.48$ & $-0.47$ & $+0.75$ & $-0.58$ & $+0.61$ & $-0.30$ \\
quote only & $+0.71$ & $-0.16$ & $+0.94$ & $-0.39$ & $+0.69$ & $-0.27$ \\
\midrule
quote + reason, 8B teacher & $+1.20$ & $-0.79$ & $+1.09$ & $-0.96$ & $+0.91$ & $-0.62$ \\
quote only, 8B teacher & $+0.82$ & $+0.21$ & $+1.17$ & $-0.22$ & $+0.90$ & $+0.03$ \\
\bottomrule
\end{tabular}}
\end{minipage}

\end{table}

\paragraph{Delivery decomposes into pointing and correcting.}
\label{sec:pointing}

A critique contains two separable components: a quote that points to the erroneous step, and a reason that explains why it is wrong. Table~\ref{tab:ablation} ablates each in turn. Localization is carried primarily by the quote: removing the reason leaves it largely unchanged, while removing the quote weakens it across tasks. Correction depends on the reason: without it, the corrective shift shrinks substantially and, with the larger teacher, can disappear or even reverse. The two components therefore play distinct roles in delivery: the quote tells the scoring teacher \emph{where} to react, while the reason tells it \emph{how} to change the student's distribution.

\section{Using the Signal at Test Time}
\label{sec:using}

We test whether the critique is useful at inference. We take wrong rollouts of
Olmo-3-32B-Think on L7-10, 55 rollouts from 39 problems, 19 of which the model
does not solve in eight samples. The model critiques each rollout with the
reference solution in context, the self-critique setting of
Table~\ref{tab:spanz}. We cut the rollout at the first sentence the critique
flags and let the model continue from the cut, eight samples per rollout. The
critique item is placed at the cut together with the cue ``Wait, I should
correct this reasoning.'' The control cuts at a fixed depth and places the cue
alone. Table~\ref{tab:tti} reports sample accuracy. With the critique in
context the model solves the problems it never solved by sampling about twice
as often as the control. The critic had the reference in context, so the gain
measures what a privileged critique delivers at inference, not unaided
self-correction. We also train with the signal (Appendix~\ref{app:training}).
On four Knights \& Knaves puzzles Olmo-3-7B-Think never solves by sampling, the
critique-conditioned arm solves two puzzles that the GRPO control never solves
(Appendix~\ref{app:cold4}). On a 320-puzzle training set, the
critique-conditioned term raises out-of-distribution accuracy over GRPO and
over reference-conditioned self-distillation, and the student as its own critic
recovers most of the oracle critic's gain (Appendix~\ref{app:kkshort}).

\begin{table}[t]
\centering
\caption{\textbf{At test time the critique's content converts math problems the
model cannot solve by sampling.} Olmo-3-32B-Think resumes its own wrong L7-10
rollouts from a cut, eight samples per rollout; sample accuracy in percent on
all problems and on the problems unsolved in eight samples. Bold = 95\% CI of
the paired difference to the control excludes 0.}
\label{tab:tti}
\small
\setlength{\tabcolsep}{6pt}
\begin{tabular}{llrr}
\toprule
cut & context at the cut & all (39) & unsolved by sampling (19) \\
\midrule
fixed depth & cue & 10.9 & 4.3 \\
critique's flagged sentence & cue + critique item & 13.4 & $\mathbf{9.3}$ \\
\bottomrule
\end{tabular}
\end{table}

\section{Conclusion and Limitations}
\label{sec:discussion}

We show that reference-conditioned self-distillation provides dense supervision but not reliable corrective credit: its signal does not localize student errors and is not informative about rollout correctness. The same teacher can recover those errors when allowed to generate a critique, suggesting that credit assignment requires an explicit search over the reasoning trace. Critique-conditioned self-distillation separates that search from scoring, using generation to identify the error and a scoring pass to turn it into dense token-level supervision. \textbf{Limitations.} Our study has three main limitations. First, our training evidence is small in scale, a 320-puzzle set and 100 steps (Appendix~\ref{app:kkshort}), and we have not run critique-conditioned self-distillation at full RL scale. Second, critique conditioning adds a generation pass over the wrong rollouts of each batch. In our runs this costs 6\% of step time with the student as critic and about a fifth with an API critic (Appendix~\ref{app:overhead}). Third, every critique in this paper, in the signal analysis, at test time and in training, is written with the reference solution in context. Our results show that a critique conveys privileged information in a form the scoring teacher can use, not that the model corrects itself unaided.

\bibliography{references}
\bibliographystyle{plainnat}

\clearpage
\appendix
\begin{center}
{\Large\bfseries Appendix}
\end{center}

\section{Training with the Critique-Conditioned Signal}
\label{app:training}
We report two training experiments. The first is a cold-start pilot on four puzzles with Olmo-3-7B-Think, the pilot referred to in \S\ref{sec:using}. The second compares the critique-conditioned term against GRPO and reference-conditioned self-distillation over 100 steps on Olmo-3-7B-Instruct, with the oracle critic and with the student as its own critic. We then report the compute cost of the critique pass.

\subsection{Cold-start pilot on four puzzles}
\label{app:cold4}

\begin{figure}[h]
\centering
\begin{minipage}[t]{0.48\linewidth}
\centering
\includegraphics[width=\linewidth]{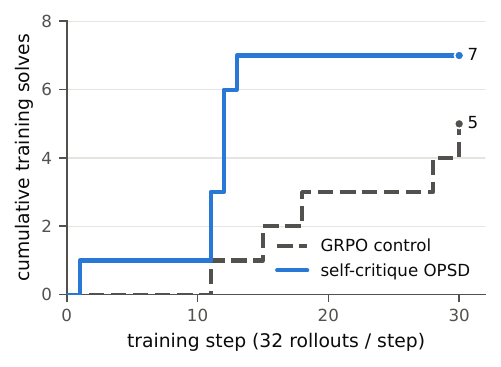}\\[-2pt]
{\small (a) all four puzzles}
\end{minipage}\hfill
\begin{minipage}[t]{0.48\linewidth}
\centering
\includegraphics[width=\linewidth]{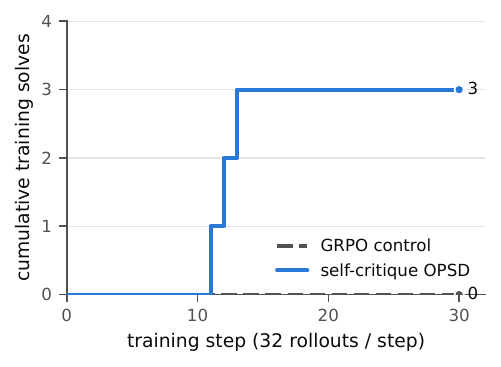}\\[-2pt]
{\small (b) without the puzzle the control solves by chance}
\end{minipage}
\caption{Cumulative solved training rollouts on four certified-cold Knights \& Knaves puzzles, critique-conditioned advantage against the GRPO control. Olmo-3-7B-Think as student, critic, and teacher.}
\label{fig:cold4}
\end{figure}

We train Olmo-3-7B-Think on four 16-person Knights \& Knaves puzzles it never
solves in 64 samples. The student writes the rollouts, critiques its own wrong
rollouts with the verified derivation as reference, and scores them as its own
teacher. On all-wrong groups, where GRPO has no gradient, the critique-conditioned
correction signal in the critic-located error span becomes the advantage, with a
zero-sum positive term on the prefix before the error. The control is GRPO alone.
Both arms run 30 steps of 32 rollouts with one seed.
Figure~\ref{fig:cold4} shows the cumulative number of solved training rollouts.
The control solves one of the four puzzles by chance and never the other three.
The critique-conditioned arm solves two of those three puzzles.

\FloatBarrier

\subsection{Training against GRPO and reference-conditioned self-distillation}
\label{app:kkshort}
We train Olmo-3-7B-Instruct \citep{olmo2026olmo3} on 320 Knights \& Knaves puzzles with 4--8 people for 100 steps, with 16 puzzles and 8 rollouts per step, learning rate $5\times10^{-6}$ and a generation cap of 8,192 tokens. We evaluate the final checkpoint at a 32,768-token cap on 500 held-out puzzles with 4--8 people and on 100 puzzles each with 9, 10 and 11 people. OOD is the mean over the three larger sizes. The baselines are GRPO (\S\ref{sec:prelim-grpo}) and reference-conditioned self-distillation with the verified derivation in the teacher's context (RLCSD, \citealp{pan2026rlcsdreinforcementlearningcontrastive}). The critique-conditioned arms add a term on the wrong rollouts. The critic reads the problem, the rollout and the derivation and writes the critique of \S\ref{sec:prelim-probe}. The teacher scores the rollout with the critique in context, and the corrective part of its log-ratio against the student is applied with weight 3 on the sentence the critique quotes and the 50 tokens after it. The critic is gpt-5-mini (oracle) or the student itself (self-critique).

\begin{table}[h]
\centering
\caption{Accuracy after 100 steps on Olmo-3-7B-Instruct. In-domain is Knights \& Knaves with 4--8 people, OOD the mean over 9, 10 and 11 people. The critique term is confined to the quoted sentence and the 50 tokens after it.}
\label{tab:kkshort}
\small
\setlength{\tabcolsep}{4pt}
\begin{tabular}{ll rrrr r}
\toprule
method & critic & in-domain & 9 & 10 & 11 & OOD \\
\midrule
base model & -- & 0.666 & 0.517 & 0.415 & 0.275 & 0.403 \\
GRPO & -- & 0.700 & 0.595 & 0.532 & 0.400 & 0.509 \\
GRPO + critique & gpt-5-mini & 0.744 & 0.680 & 0.568 & 0.445 & 0.564 \\
\midrule
RLCSD & -- & 0.745 & 0.688 & 0.583 & 0.430 & 0.567 \\
RLCSD + critique & gpt-5-mini & 0.754 & 0.698 & 0.623 & 0.497 & 0.606 \\
RLCSD + critique & self & 0.735 & 0.693 & 0.595 & 0.490 & 0.593 \\
\bottomrule
\end{tabular}
\end{table}

The critique term on top of GRPO is worth as much as the reference-conditioned teacher term, +5.5 against +5.8 OOD points over GRPO, and on top of RLCSD it adds a further 3.9 (Table~\ref{tab:kkshort}). The two gains add. The self-critique is the weaker critic, its quotes match a sentence of the rollout 82\% of the time against 96\% for the oracle, and it still recovers most of the oracle's gain, 0.593 against 0.606 with RLCSD at 0.567.

\subsection{Compute cost of the critique pass}
\label{app:overhead}
Critique conditioning adds one generation per critiqued rollout. Only wrong rollouts in groups with at least one correct rollout are critiqued, 16 to 18 of the 128 rollouts per step in the runs of Table~\ref{tab:kkshort}. Table~\ref{tab:overhead} gives the cost per step averaged over the 100 steps. The self-critique runs in the same inference engine as the rollouts. We push the frozen base weights, generate the critiques and push the current policy back. Olmo-3-7B-Instruct writes short critiques, 175 tokens on average, so the critique tokens are 0.4\% of the rollout tokens generated per step and the whole pass is 6\% of step time, one point of which is the weight swap. The API critic is a reasoning model and returns about 3,000 tokens per critique including its reasoning. Its share of step time is time spent waiting on the API, not GPU time, and could overlap with rollout generation.

\begin{table}[h]
\centering
\caption{Cost of the critique pass per training step, mean over 100 steps, in the regime of Table~\ref{tab:kkshort}. Tokens are per critique.}
\label{tab:overhead}
\footnotesize
\setlength{\tabcolsep}{3.5pt}
\begin{tabular}{llrrrrr}
\toprule
critic & arm & critiqued rollouts & prompt tokens & completion tokens & step time & share \\
\midrule
gpt-5-mini (API) & RLCSD + critique & 16 & 5,600 & 2,970 & 255 s & 21\% \\
gpt-5-mini (API) & GRPO + critique & 18 & 4,960 & 2,890 & 256 s & 22\% \\
self (in-engine) & RLCSD + critique & 18 & 6,770 & 175 & 219 s & 6\% \\
\bottomrule
\end{tabular}
\end{table}
\FloatBarrier

\section{Critique Examples}
\label{app:examples}
We show one wrong Qwen3-1.7B rollout per task with its gold error sentence, the gpt-5.2 critique and the student's own critique. Both critics received the same prompt (Appendix~\ref{app:prompt-critique}) and the same reference. The gold error sentence is in italics in the rollout excerpt.

\paragraph{Knights \& Knaves.}
\begin{small}
\noindent\textbf{Problem.} A very special island is inhabited only by knights and knaves. Knights always tell the truth, and knaves always lie. You meet 6 inhabitants on this island: Yara, Liam, Peter, Sophia, Noah, and Bob. Yara says: ``Peter is a knave if and only if Liam is a knight.'' Liam says: ``It is not the case that Yara is a knave if and only if Yara is a knight.'' Peter says: ``Bob is a knight.'' Sophia says: ``Liam is a knight.'' Noah says: ``It is not the case that Liam is a knave.'' Bob says: ``Noah is a knave.''

\noindent\textbf{Reference answer.} Yara is a knight, Liam is a knight, Peter is a knave, Sophia is a knight, Noah is a knight, Bob is a knave.

\noindent\textbf{Student's answer.} Yara is a knight, Liam is a knave, Peter is a knight, Sophia is a knave, Noah is a knave, Bob is a knight.

\noindent\textbf{Rollout excerpt.} [\ldots] if Yara is a knight ($Y$ is true), then the statement ``Yara is a knave if and only if Yara is a knight'' would be ``False $\leftrightarrow$ True'', which is False. Similarly, if Yara is a knave ($Y$ is false), then the statement is ``True $\leftrightarrow$ False'', which is also False. Therefore, the statement is always false. \emph{Therefore, Liam's statement is false.} Therefore, if Liam is a knight, he would be making a false statement, which is impossible. Therefore, Liam must be a knave. Wait, that seems important. [\ldots]

\noindent\textbf{Gold error.} Liam's statement is the negation of the inner biconditional, and since that inner biconditional is always false, Liam's statement is actually true.

\noindent\textbf{Oracle critique (gpt-5.2).}
ERROR 1: ``Therefore, the statement is always false.'' -- The biconditional ``Yara is a knave iff Yara is a knight'' is always false, so its negation (Liam's statement) is always true, not false.\\
ERROR 2: ``Therefore, Liam must be a knave.'' -- Since Liam's statement is necessarily true, Liam must be a knight, not a knave.\\
ERROR 3: ``Since we just determined that Liam is a knave,'' -- This propagates the earlier incorrect classification of Liam, making subsequent deductions unreliable.

\noindent\textbf{Self-critique (Qwen3-1.7B).}
ERROR 1: ``Liam is a knave'' -- Liam's statement is a contradiction (always false), so he must be a knave. This is correct, but the final answer contradicts this conclusion.\\
ERROR 2: ``Noah is a knave'' -- Noah's statement is ``Liam is a knight,'' which is false because Liam is a knave. Therefore, Noah is a knave. This is correct.\\
ERROR 3: ``Bob is a knight'' -- Bob's statement ``Noah is a knave'' is true, so Bob is a knight. This is correct.\\
ERROR 4: ``Peter is a knight'' -- Peter's statement ``Bob is a knight'' is true, so Peter is a knight. This is correct.\\
ERROR 5: ``Yara is a knight'' -- Yara's statement is true, so she is a knight. This is correct.\\
Final Answer (boxed): Yara is a knight, Liam is a knight, Peter is a knave, Sophia is a knight, Noah is a knight, Bob is a knave.
\end{small}

\noindent The oracle quotes the sentence before the gold error and the conclusion drawn from it, and each reason says which way to correct it. The self-critique quotes fragments of the final answer, endorses the student's reasoning, and then writes the reference answer as its own final answer. Its quotes carry no location and its reasons carry no correction before that last line.

\paragraph{DeepMath L5-7.}
\begin{small}
\noindent\textbf{Problem.} Find the residue of the function $\sin\left(\frac{1}{z}\right)$ at its poles.

\noindent\textbf{Reference answer.} 1. \quad \textbf{Student's answer.} 0.

\noindent\textbf{Rollout excerpt.} [\ldots] So, this is an infinite series, and there are infinitely many terms with positive powers of $1/z$. Therefore, the Laurent series has an infinite number of terms with negative powers of $z$, which means that $z=0$ is an essential singularity, not a pole. But wait, the question says ``at its poles''. \emph{But if there are no poles, then the residue would be\ldots\ Well, if there are no poles, then the residue is zero?} But maybe I need to check if there are any poles. But since $\sin(1/z)$ has an essential singularity at $z=0$, there are no poles. So, the residue at the pole\ldots\ but there are no poles. Therefore, the residue is zero? [\ldots]

\noindent\textbf{Gold errors.} Three sentences are gold. The first is the italic one above (a residue can exist at an essential singularity and is the coefficient of the $z^{-1}$ term, which is nonzero here). The other two are ``Since there are no poles, the Laurent expansion does not have any $(z-a)^{-1}$ term, so the coefficient is zero'' and ``if a function has an essential singularity at a point, then the residue is not defined in the usual sense.''

\noindent\textbf{Oracle critique (gpt-5.2).}
ERROR 1: ``then the residue would be\ldots\ Well, if there are no poles, then the residue is zero?'' -- Lack of poles does not imply the residue is zero; residue is defined at isolated singularities (including essential ones) via the $z^{-1}$ coefficient.\\
ERROR 2: ``if a function has an essential singularity at a point, then the residue is not defined in the usual sense.'' -- False: the residue is still defined for essential singularities as the coefficient of $z^{-1}$ in the Laurent series.\\
ERROR 3: ``Since there are no poles, the Laurent expansion does not have any $(z-a)^{-1}$ term'' -- Incorrect: $\sin(1/z)=1/z-1/(6z^3)+\cdots$ does contain a $z^{-1}$ term, so residue at $0$ is $1$.

\noindent\textbf{Self-critique (Qwen3-1.7B).}
ERROR 1: ``the residue of the function $\sin\left(\frac{1}{z}\right)$ at its poles is $\boxed{0}$'' -- The student incorrectly concludes that the residue is zero because there are no poles, but the residue is defined only for isolated singularities, not for the absence of poles. The function has an essential singularity at $z=0$, and the residue at that singularity is 1, not zero.\\
ERROR 2: ``the student's final answer is known to be WRONG: the solution below contains at least one mistake'' -- The student's final answer is wrong because the residue of $\sin\left(\frac{1}{z}\right)$ at its essential singularity (not poles) is 1, not 0.\\
{}[ERROR 3 to ERROR 5 repeat these two items.]
\end{small}

\noindent The oracle's three quotes are the three gold error sentences, and each reason states the correct fact. The self-critique quotes the final answer and a line of its own prompt, neither of which is a step of the derivation, so it does not localize. Its reason is right: the residue at the essential singularity is 1. This is the division of labour of Table~\ref{tab:ablation}. The quote carries the location and the reason carries the direction, and here the self-critique has the second without the first.

\section{Extended Related Work}
\label{app:related}
\textbf{Self-distillation with a privileged teacher.} On-policy
self-distillation
\citep{zhao2026selfdistilledreasoneronpolicyselfdistillation,hubotter2026reinforcementlearningselfdistillation}
produces a per-token signal in place of the single bit that
verifiable-reward RL provides: the same model scores the student's own
rollout with the reference solution in its context. The line has grown by
varying pieces of that recipe. RLSD keeps the reference fixed and turns the
log-ratio into a token-level advantage weight
\citep{yang2026selfdistilledrlvr}. RLCSD contrasts a correct hint against a
wrong one to suppress the style drift a hint alone induces
\citep{pan2026rlcsdreinforcementlearningcontrastive}, and
\citet{zhao2026privilegedinformationbettersolution} replace the full
solution with a more structured trace. Carried over to thinking models, the
recipe degrades performance rather than helping
\citep{kaur2026rethinkingonpolicyselfdistillationthinking}, and Rebellious
Student reverses the signal on rollouts the student already gets right, to
encourage exploration rather than correction
\citep{kim2026rebelliousstudentreversingteacher}. Closest to our premise,
\citet{li2026localizingcreditdivergencepathconditioned} show that a teacher
conditioned only on the final answer produces no signal at the intermediate
positions where guidance is needed, and condition it on a successful peer
rollout instead. All of these variants take the scored signal as given and
change how it is consumed or what the teacher is conditioned on. We ask
whether the scored signal carries any credit at the error, and find that a
full reference solution does not. Cross-model distillation selects the
teacher's tokens by mismatch, importance, learnability, position, or
evidence
\citep{yu2026mismatchmattersonpolicydistillation,xu2026tiptokenimportanceonpolicy,wang2026disagreementlearnabletokenteachability,liu2026teachertokensreliablepositionweighted,xiao2026findingevidencediscoveringdecisionsupporting},
and \citet{li2026rethinkingonpolicydistillationlarge} find instead that the
high-overlap tokens carry the signal. Those methods assume a signal to
select. A cross-model teacher's gap reflects a real capability difference;
a privileged teacher differs from the student only by context, and at the
errors we find no signal to select.

\textbf{Failure modes of the privileged teacher.} Three papers report that
the privileged teacher misbehaves.
\citet{harne2026privilegedbiasedpiconditionedteachers} show that the teacher
is biased by the privileged information it is given, and ask whether
self-distillation alone teaches anything.
\citet{kaur2026rethinkingonpolicyselfdistillationthinking} show that the
same information degrades performance on thinking models.
\citet{pan2026rlcsdreinforcementlearningcontrastive} find that the
distributional gap concentrates on style tokens rather than task-bearing
ones. None of them anchors the diagnosis to the place where the student's
reasoning goes wrong. We measure the signal itself against gold error spans
on the student's own rollouts, and find that scoring fails to localize the
error while generating does.

\textbf{Locating an error is a reasoning problem.}
\citet{tyen2024llmsreasoningerrorscorrect} show that language models cannot
find their reasoning errors but can correct them given the error location,
and \citet{huang2024largelanguagemodelsselfcorrect} and
\citet{kamoi2024can} show that intrinsic
self-correction without an external signal does not improve reasoning.
Verifiers do better when they reason before they judge. Generative
verifiers \citep{zhang2025generativeverifiersrewardmodeling},
critique-out-loud reward models
\citep{ankner2024critiqueoutloudrewardmodels}, and reward models trained as
reasoners \citep{chen2026rmr1rewardmodelingreasoning} all take this route,
and process reward models that think locate erroneous steps better than
discriminative scorers with a fraction of the labels
\citep{khalifa2025processrewardmodelsthink}. ProcessBench reports that
trained step scorers underperform prompted critics
\citep{zheng2025processbenchidentifyingprocesserrors}, and
\citet{chen2026learningselfverifymakeslanguage} find that verification is a
capability separate from generation. Generation's advantage is not
universal. \citet{tyen2024llmsreasoningerrorscorrect} also show that a
trained classifier can localize mistakes without generating, and
\citet{stechly2023gpt4doesntknowits} find a critique's content irrelevant to
the benefit of iterative prompting in planning. Our claim is
correspondingly narrow. The token-level likelihoods of a teacher scoring a
forced trace do not localize the error, while the same model generating a
critique does.

\textbf{Process supervision and critics.} Process supervision identifies
the error when the step-level labels are created. Labels come from humans
\citep{uesato2022solvingmathwordproblems,lightman2023letsverifystepstep},
from automatic rollouts \citep{wang2024mathshepherdverifyreinforcellms}, or
from an LLM judge, and the automatically estimated labels generalize worse
than the judged or human ones \citep{zhang2025lessonsdevelopingprocessreward}.
The generative verifiers and critique-out-loud models above, together with
critics trained to critique rather than imitate
\citep{wang2025critiquefinetuninglearningcritique}, put the search into
generation, which is the readout we use. Such critics are scored against
human-annotated step-level error benchmarks
\citep{tyen2024llmsreasoningerrorscorrect,zheng2025processbenchidentifyingprocesserrors,he2025largelanguagemodelsdetect},
which we revisit in Appendix~\ref{app:human}. Critique-conditioned
self-distillation combines both ideas: a generated critique, delivered
through the scoring teacher, becomes a per-token signal for on-policy
self-distillation, with the critic and the teacher the same model.

\section{Setup Details}
\label{app:setup}
\paragraph{Rollouts and targets.}
Every rollout in the paper is sampled from the student itself, on the
training distribution. A rollout is a \emph{target} if the verifier scores it
wrong, it is not truncated, and it ends in a boxed answer; verifier-correct
rollouts of the same problems are kept as a control stratum for
\S\ref{sec:predictive}. K\&K targets are drawn only from puzzles whose English
rendering is decidable as worded, about half of the pool, because on the
remainder a ``wrong'' rollout is often a legitimate alternative reading.
Rollouts are sampled with the training parameters, temperature 1.0, top-p
0.95, top-k 20, eight rollouts per prompt, with a 16{,}384-token budget for
Qwen3-1.7B and a 32{,}768-token budget for the Olmo-3 models, whose thinking
traces are longer.

\paragraph{Gold error spans.}
A grader model reads the problem, the rollout, and the reference, and lists
the places where the reasoning goes wrong, earliest first, each as a verbatim
quote of one sentence with a reason, a severity, and a confidence
(\S\ref{app:prompt-grader}). The grader is deepseek-reasoner for the original
Qwen sample, deepseek-v4-flash for the problems added in the scale-up, and
claude-sonnet-5 for the two Olmo-3 studies. Each flagged sentence is then judged
independently by gpt-5.2 and by claude-sonnet-5, which see the problem, the
reference, and the rollout with the sentence marked and the grader's reason
attached (\S\ref{app:prompt-jury}). A quarter of the flags sent to each jury
are replaced by position-matched control sentences carrying the same reason,
which measures the jury's leniency; the control acceptance rate is reported
with each study. A flag is \emph{strict} gold if both juries accept it and
\emph{lenient} gold if the claude jury accepts it. Error-span Z uses the first
strict gold sentence of each rollout; the curves of Figs.~\ref{fig:localization}
and \ref{fig:critique} use every strict gold sentence; the critique gate of
Table~\ref{tab:gate} uses the lenient tier. In the Olmo-3 studies the claude jury
shares its model with the grader, so gpt-5.2 is the only fully independent
judge there.

\paragraph{Placebos.}
Every conditioning block is scored against a placebo of the same shape, and
all reported signals are block minus placebo. The placebo for the reference is
a different problem's reference of matched length, assigned by one fixed
derangement of the problems that every arm shares. The placebo for a critique
is the critique written for a different problem's rollout, assigned by the
same derangement. The undifferenced arms are reported as negative controls in
Table~\ref{tab:spanz-controls}.

\paragraph{Error-span Z windows and statistics.}
The error window is the 20 tokens immediately after the first gold error
sentence ends. Control windows have the same shape and sit after the ends of
clean sentences before the first error. We use the 12 nearest control windows,
non-overlapping, with a 60-token guard band around the error sentence. On the
localization signal we take the log of each window mean before Z-scoring; the
correction signal is already a signed log-probability difference and is used as
is. The dataset-level statistic is the mean of $Z_r$ over rollouts, winsorized
at the 5th and 95th percentiles, with a 95\% bootstrap interval clustered by
problem (2{,}000 resamples). The null subtracted from every cell relabels a
control sentence of the same rollout as the error and recomputes $Z_r$; the
reported value is $Z - Z_{\mathrm{null}}$. Bold in Table~\ref{tab:spanz} marks
cells whose 95\% interval excludes zero.

\paragraph{Which reference each model sees.}
The scoring teacher's reference is a step-by-step derivation verified against
the ground truth for K\&K and the dataset's reference solution with its final
answer for DeepMath. The oracle critic receives the same reference as the
teacher. The open-weights critics on the scale-up sample receive the K\&K
derivations and, for DeepMath, a clean derivation generated and verified
against the ground-truth answer, because the dataset solutions are raw
chains of thought that weaker critics mistake for a second student solution;
the 60 rollouts without a verified derivation are dropped from those arms.

\subsection{Critique prompt}
\label{app:prompt-critique}
The critic receives the problem, the student's rollout with sentence tags, the
reference solution, and the instructions below. The rollout is presented
before the reference to prevent the critic from quoting the reference. The
same prompt is used for every critic.
\begin{small}
\begin{verbatim}
Problem: {problem}

You are grading a student's solution to this problem.

The student's final answer is known to be WRONG: the solution below
contains at least one mistake.

=== Student Solution Begin ===
[S0] <sentence 0> [S1] <sentence 1> ...
=== Student Solution End ===

For reference, here is a correct solution:

=== Reference Solution Begin ===
{reference}
=== Reference Solution End ===

Coverage matters more than precision here: if you are unsure where the
reasoning FIRST goes wrong, list EACH plausible location as its own
ERROR line (up to 5), earliest first -- it is better to list several
candidate locations than to miss the true one.

Write a CONCISE critique of the student's reasoning, at most 120 words
total. Use EXACTLY this format, one line per error, earliest error
first, at most 5 errors, and nothing else:

ERROR 1: "<short quote, at most 10 words, copied word-for-word from the
student's solution at the point the error occurs>" -- <one short
sentence: what is wrong there>
ERROR 2: ...

If the reasoning contains no errors, output the single line: NO ERRORS

Rules: each quote must be copied exactly, character for character, from
the Student Solution above (never from the reference, never a
paraphrase), and must be at most 10 words. Point at the sentence where
the reasoning first goes wrong, not merely the final answer line. Only
report NEW errors: once the reasoning is broken, list a later step only
if it introduces a NEW, independent error. Report only errors you are
CERTAIN about.

IMPORTANT REMINDER: every quote must be copied word-for-word from
between "=== Student Solution Begin ===" and "=== Student Solution End
===" above -- NEVER from the Reference Solution, which is correct and is
not being graded.
The student's sentences are tagged [S0], [S1], ... to help you keep
track of position. Do NOT include any [S#] tag inside your quotes --
quote only the sentence text itself.
\end{verbatim}
\end{small}

\subsection{Gold-span grader prompt}
\label{app:prompt-grader}
The grader sees the problem, the rollout, and the reference (arm \emph{ref}),
with the same known-wrong line as the critic, followed by:
\begin{small}
\begin{verbatim}
List the places where the student's REASONING goes wrong, in order of
occurrence, starting with the FIRST error. Report ONLY errors you are
CERTAIN about -- one certain error is worth more than several doubtful
ones. List at most 8. Your final answer must use EXACTLY this format and
nothing else, beginning with the line "ERROR 1:" (or the single line
"NO ERRORS" if you find none):

ERROR 1:
QUOTE: "<one sentence copied word-for-word from the student's solution,
at the point where the error occurs>"
REASON: <one short sentence explaining what is wrong there>
SEVERITY: <integer 1-5: how much this error contributes to a wrong final
answer; 5 = directly causes it, 1 = minor slip>
CONFIDENCE: <integer 0-100: how certain you are that this is a genuine
error>

ERROR 2:
...

Rules: each QUOTE must be copied exactly, character for character, from
the Student Solution above -- not from the reference, and not your own
paraphrase. Point to the sentence in the student's reasoning where
things go wrong, not merely the final answer line. ERROR 1 must be the
EARLIEST point where the reasoning actually goes wrong. Only report NEW
errors: once the reasoning is broken, do not list every later step that
merely inherits an earlier mistake -- list a later step only if it
introduces a NEW, independent error. Do not include errors you are not
sure about.
\end{verbatim}
\end{small}

\subsection{Jury prompt}
\label{app:prompt-jury}
Each grader flag is judged independently by two models. The judge sees the
problem, the reference, and the rollout with the flagged sentence marked
between \texttt{<<<} and \texttt{>>>}; a quarter of the flags are replaced by
position-matched control sentences with the grader's reason attached
unchanged.
\begin{small}
\begin{verbatim}
Above is a problem, a correct reference solution, and a student's
solution in which one sentence has been marked between <<< and >>>. A
grader claims the student's reasoning goes wrong AT THAT SENTENCE, for
this reason:

GRADER'S CLAIM: {reason}

Judge the claim on the student's actual reasoning. Answer YES only if a
genuine reasoning error is committed at (or within a sentence or two of)
the marked sentence -- i.e. the marked step is wrong given what the
student derived so far, not merely downstream of an earlier mistake, and
not a legitimate alternative reading. Answer NO otherwise.

Reply in EXACTLY this format:
VERDICT: <YES or NO>
CONFIDENCE: <integer 0-100>
WHY: <one short sentence>
\end{verbatim}
\end{small}

\section{Error-Span Z: Full Results}
\label{app:spanz}
This section reports error-span Z with 95\% confidence intervals for every
conditioning and teacher on the Qwen3-1.7B rollouts, including the 8B model as
a scale-control teacher and as a critic, together with the undifferenced hint
arms as negative controls and the per-rollout sign consistency of the
oracle-critique signal. Intervals are cluster bootstraps by problem. The corresponding tables for the Olmo-3 students are in
Appendix~\ref{app:olmo}.

\begin{table}[h]
\centering
\caption{Error-span Z at the first gold error with 95\% confidence intervals, Qwen3-1.7B rollouts, scale-up sample, every conditioning and teacher. Rows: the reference solution or a critique placed in the scoring teacher's context; teacher Qwen3-1.7B (the student itself) or Qwen3-8B; critiques by gpt-5.2 (oracle), by the 1.7B student on its own rollouts, or by Qwen3-8B. Localization on log content KL (positive = the distribution reacts at the error); correction on $\dlogp$ of the student's token (negative = corrective). $n$ = rollouts entering the cell.}
\label{tab:spanz-full}
\scriptsize
\setlength{\tabcolsep}{3pt}
\begin{tabular}{lllrlrlr}
\toprule
conditioning & teacher & task & loc & 95\% CI & corr & 95\% CI & $n$ \\
\midrule
reference & 1.7B & K\&K & $-0.17$ & $[-0.33,-0.01]$ & $+0.11$ & $[-0.04,+0.26]$ & 317 \\
 & & L5-7 & $+0.14$ & $[-0.01,+0.29]$ & $-0.14$ & $[-0.30,+0.02]$ & 430 \\
 & & L7-10 & $+0.07$ & $[-0.08,+0.19]$ & $+0.17$ & $[+0.07,+0.28]$ & 690 \\
reference & 8B & K\&K & $+0.13$ & $[-0.03,+0.29]$ & $-0.16$ & $[-0.35,+0.02]$ & 317 \\
 & & L5-7 & $+0.05$ & $[-0.11,+0.20]$ & $-0.06$ & $[-0.23,+0.10]$ & 430 \\
 & & L7-10 & $-0.05$ & $[-0.18,+0.08]$ & $+0.06$ & $[-0.06,+0.19]$ & 690 \\
\midrule
critique, oracle & 1.7B & K\&K & $+0.75$ & $[+0.54,+0.97]$ & $-0.62$ & $[-0.86,-0.40]$ & 317 \\
 & & L5-7 & $+0.99$ & $[+0.82,+1.16]$ & $-0.90$ & $[-1.14,-0.65]$ & 448 \\
 & & L7-10 & $+0.78$ & $[+0.63,+0.92]$ & $-0.54$ & $[-0.71,-0.35]$ & 711 \\
critique, oracle & 8B & K\&K & $+1.20$ & $[+0.95,+1.45]$ & $-0.79$ & $[-1.10,-0.49]$ & 317 \\
 & & L5-7 & $+1.09$ & $[+0.91,+1.27]$ & $-0.96$ & $[-1.25,-0.68]$ & 448 \\
 & & L7-10 & $+0.91$ & $[+0.77,+1.05]$ & $-0.62$ & $[-0.79,-0.45]$ & 711 \\
\midrule
critique, 1.7B self & 1.7B & K\&K & $+0.16$ & $[-0.01,+0.32]$ & $-0.07$ & $[-0.23,+0.11]$ & 316 \\
 & & L5-7 & $+0.25$ & $[+0.10,+0.40]$ & $-0.14$ & $[-0.29,+0.01]$ & 438 \\
 & & L7-10 & $+0.18$ & $[+0.06,+0.30]$ & $-0.07$ & $[-0.19,+0.04]$ & 648 \\
critique, 8B critic & 1.7B & K\&K & $+0.22$ & $[+0.03,+0.39]$ & $-0.30$ & $[-0.50,-0.10]$ & 317 \\
 & & L5-7 & $+0.20$ & $[+0.05,+0.34]$ & $-0.19$ & $[-0.34,-0.05]$ & 427 \\
 & & L7-10 & $+0.25$ & $[+0.12,+0.37]$ & $+0.03$ & $[-0.08,+0.14]$ & 634 \\
\bottomrule
\end{tabular}
\end{table}

\begin{table}[h]
\centering
\caption{Negative controls and reliability, Qwen3-1.7B rollouts, scale-up sample. Left: error-span Z of each privileged context alone, without subtracting its placebo (reference alone; a donor critique alone), which should sit at zero. Right: per-rollout sign consistency of the oracle-critique signal, the share of rollouts whose first error window departs from its own control windows in the expected direction, on each channel. $^*$: 95\% CI excludes 0.}
\label{tab:spanz-controls}
\scriptsize
\begin{tabular}{llrrrr}
\toprule
task & teacher & reference alone & critique placebo alone & sign\% loc & sign\% corr \\
\midrule
K\&K & 1.7B & $-0.13^*$ & $-0.06$ & 68.1 & 55.5 \\
K\&K & 8B & $-0.15^*$ & $-0.08$ & 72.2 & 55.8 \\
L5-7 & 1.7B & $-0.15$ & $-0.07$ & 74.8 & 55.6 \\
L5-7 & 8B & $-0.12$ & $-0.14$ & 74.8 & 56.7 \\
L7-10 & 1.7B & $-0.27^*$ & $-0.29^*$ & 67.1 & 46.0 \\
L7-10 & 8B & $-0.25^*$ & $-0.27^*$ & 68.5 & 49.6 \\
\bottomrule
\end{tabular}
\end{table}

\section{Full Figures}
\label{app:figs}
This section reports the figures of the main text for the remaining
settings: all three tasks including DeepMath 7-10 at both alignments, the
Qwen3-8B model as the scoring teacher, and the student's own critiques in place
of the oracle's. Across every setting the pattern of the main text holds: with
the reference in context the error and control curves move together, and with
a critique in context the signal rises at the error sentences and nowhere
else, more strongly for the 8B teacher and more weakly for the student's own
critiques.

\begin{figure}[h]
\centering
\includegraphics[width=\linewidth]{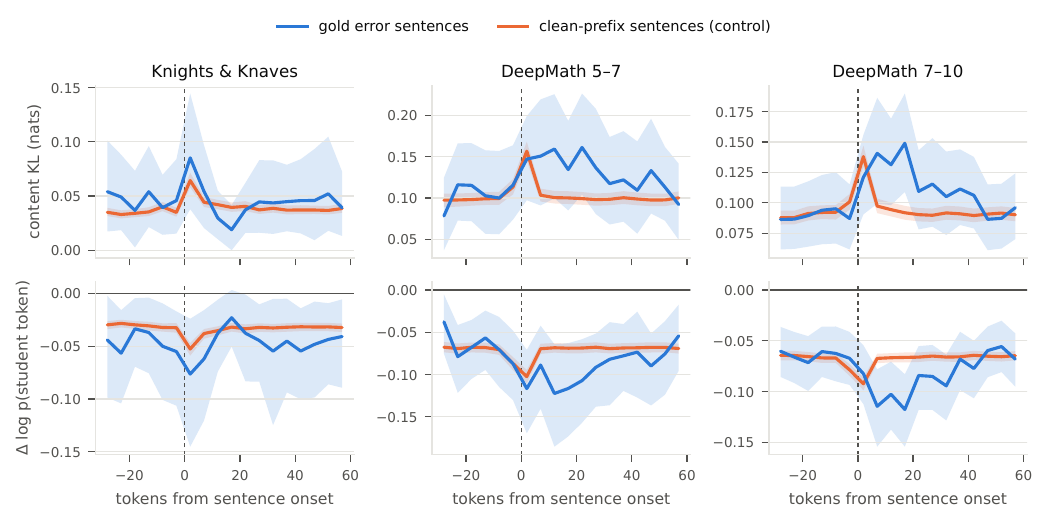}\\[2pt]
\includegraphics[width=\linewidth]{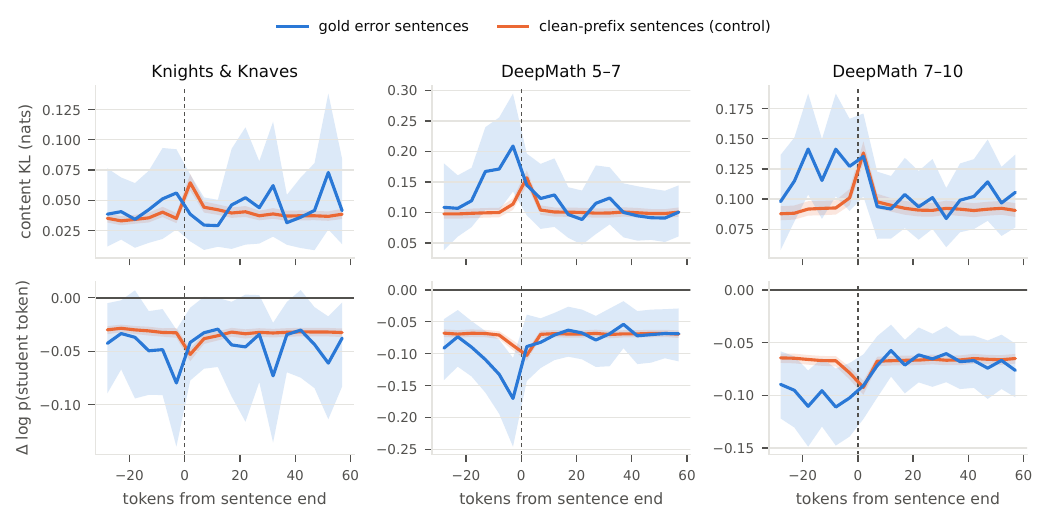}
\caption{Reference-conditioned signal on all three tasks. Content KL (top row of each block) and $\dlogp$ of the student's token (bottom row) around gold error sentences (blue) and clean-prefix sentences (orange), aligned to the sentence onset (upper block) and to the sentence end (lower block). Qwen3-1.7B scoring its own rollouts with the reference solution in context, minus the shuffled-reference placebo; scale-up sample.}
\label{fig:app-loc-3task}
\end{figure}
\begin{figure}[h]
\centering
\includegraphics[width=\linewidth]{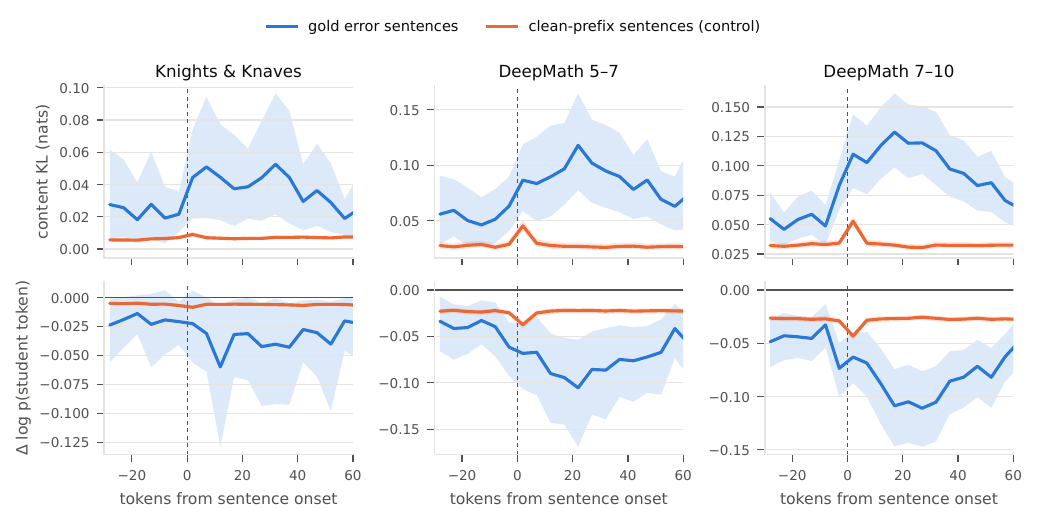}\\[2pt]
\includegraphics[width=\linewidth]{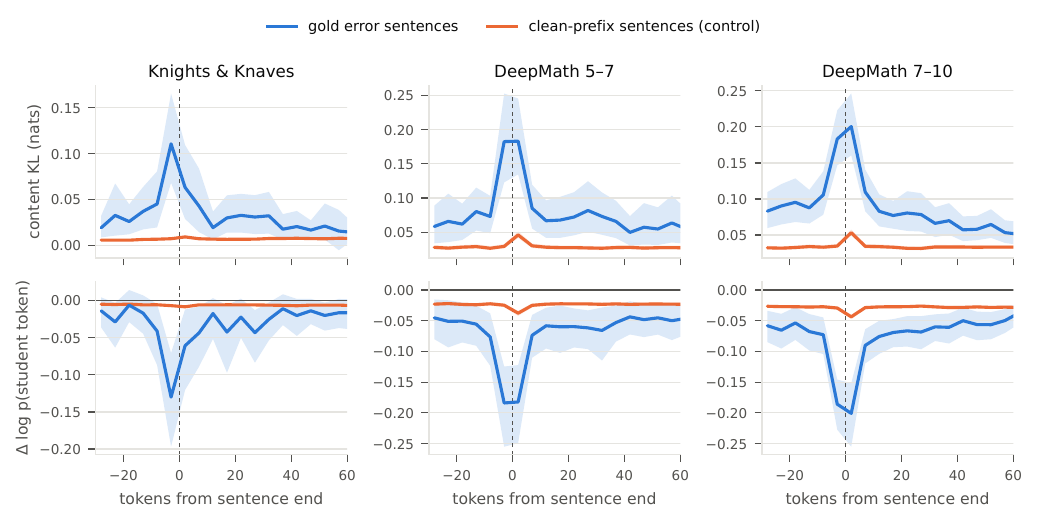}
\caption{Critique-conditioned signal on all three tasks. Same quantities, alignments, and rollouts as the figure above, with gpt-5.2's critique of the rollout in the teacher's context in place of the reference, minus the donor-critique placebo. Teacher Qwen3-1.7B.}
\label{fig:app-crit-3task}
\end{figure}
\begin{figure}[h]
\centering
\includegraphics[width=\linewidth]{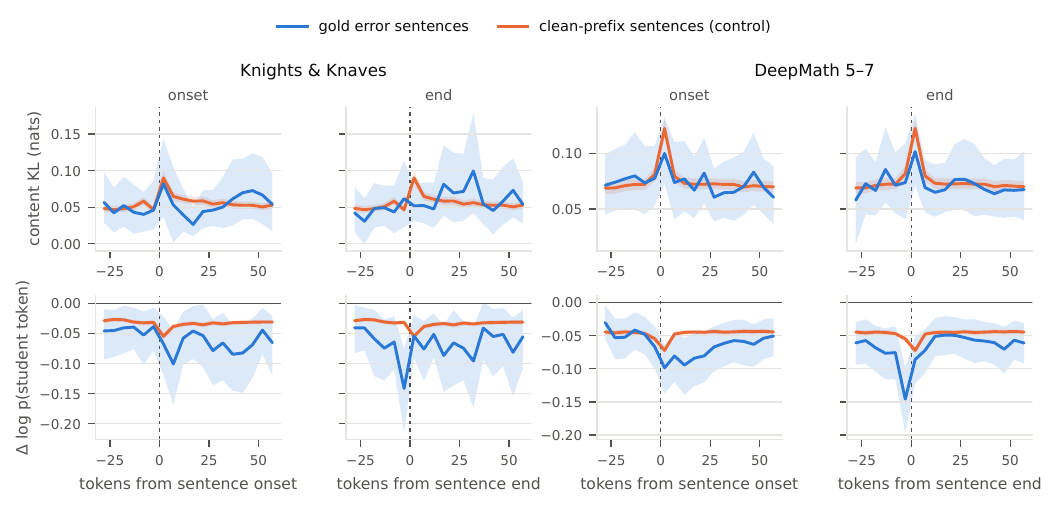}
\caption{Reference-conditioned signal with Qwen3-8B as the scoring teacher on the Qwen3-1.7B rollouts. Layout as Fig.~\ref{fig:localization}: content KL (top) and $\dlogp$ of the student's token (bottom), Knights \& Knaves and DeepMath 5-7, onset- and end-aligned; reference minus shuffled-reference placebo; scale-up sample.}
\label{fig:app-twin-8b}
\end{figure}
\begin{figure}[h]
\centering
\includegraphics[width=\linewidth]{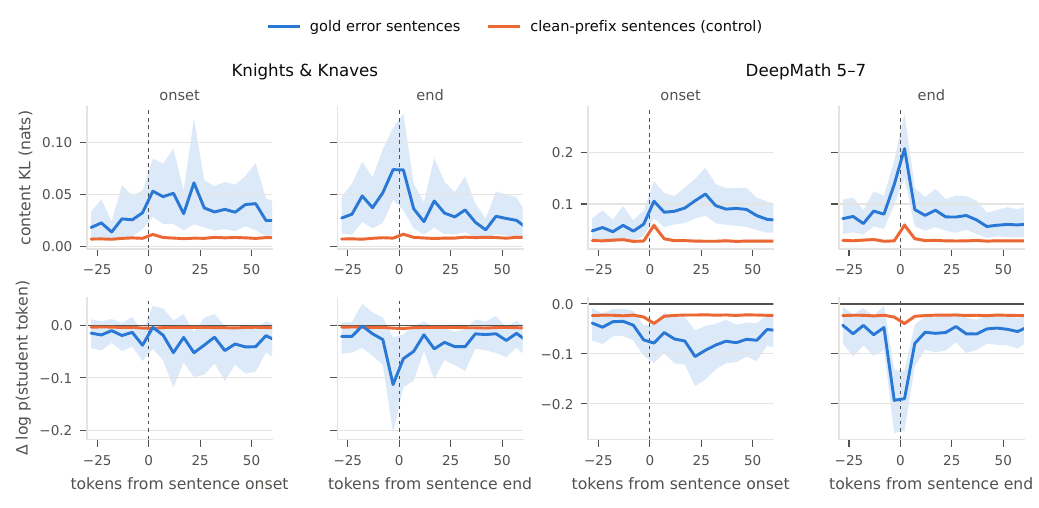}
\caption{Critique-conditioned signal with Qwen3-8B as the scoring teacher on the Qwen3-1.7B rollouts, gpt-5.2 critiques. Layout as Fig.~\ref{fig:critique}: content KL (top) and $\dlogp$ of the student's token (bottom), Knights \& Knaves and DeepMath 5-7, onset- and end-aligned; critique minus donor-critique placebo; scale-up sample.}
\label{fig:app-crit-8b}
\end{figure}

\begin{figure}[h]
\centering
\includegraphics[width=\linewidth]{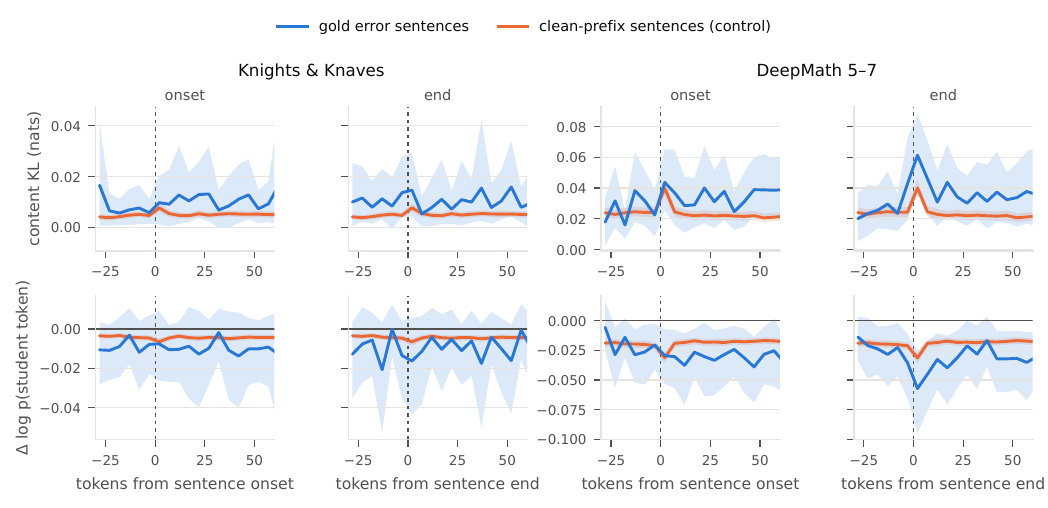}
\caption{Critique-conditioned signal with the student's own critiques: Qwen3-1.7B writes the critique, then scores its own rollout with that critique in context. Layout as Fig.~\ref{fig:critique}: content KL (top) and $\dlogp$ of the student's token (bottom), Knights \& Knaves and DeepMath 5-7, onset- and end-aligned; critique minus donor-critique placebo; scale-up sample. Note the smaller vertical scale.}
\label{fig:app-crit-self}
\end{figure}

\section{Correctness Prediction: Full Results}
\label{app:pred}
This section gives the design and the full numbers behind \S\ref{sec:predictive}.
The primary test is within-problem: for every problem that has at least one
wrong and at least one correct rollout, 34 / 28 / 49 problems on K\&K /
DeepMath 5-7 / DeepMath 7-10, each with exactly two correct siblings, we ask
whether the trace-mean of a signal orders the correct rollouts above the wrong
ones, and report the pairwise AUC with a 95\% interval from 2{,}000 bootstrap
draws clustered by problem (Table~\ref{tab:pred-within}). The pooled view, in
which all correct rollouts of a task are scored against all wrong ones, is
Fig.~\ref{fig:predictive}; it is shown because the content-free placebos
reproduce its deviations from chance, which is what identifies them as length
and difficulty effects. The trained probe (Table~\ref{tab:pred-probe}) is an
L2 logistic regression on about sixty per-trace summaries of each signal,
moments, quantiles, spike rates, and positional bins, fitted with grouped
cross-validation by problem and scored out of fold; the length baseline uses
the response length and its logarithm only, and the length of the critique is
excluded from the critique arms because it encodes the critique's own verdict.
K\&K is the weak task for this design, since wrong rollouts run much longer
there and length alone reaches a pooled AUC of 0.71; the DeepMath splits are
the clean case. One comparison in Table~\ref{tab:pred-probe} is worth
isolating: the full reference, which contains the answer, predicts correctness
worse than the bare answer at both scales. All numbers in this section are on
the original sample.

\begin{table}[h]
\centering
\caption{Within-problem pairwise AUC of the reference-conditioned trace-mean signal, correct vs.\ wrong sibling rollouts of the same problem, with 95\% CI. 0.5 = chance; no cell excludes it.}
\label{tab:pred-within}
\scriptsize
\begin{tabular}{llll}
\toprule
task & teacher & trace-mean KL & trace-mean $\dlogp$ \\
\midrule
K\&K & 1.7B & 0.598 [0.451, 0.740] & 0.396 [0.260, 0.536] \\
K\&K & 8B & 0.435 [0.295, 0.582] & 0.630 [0.490, 0.759] \\
L5-7 & 1.7B & 0.609 [0.448, 0.780] & 0.487 [0.329, 0.634] \\
L5-7 & 8B & 0.495 [0.341, 0.668] & 0.529 [0.364, 0.686] \\
L7-10 & 1.7B & 0.602 [0.490, 0.709] & 0.425 [0.317, 0.527] \\
L7-10 & 8B & 0.566 [0.456, 0.679] & 0.481 [0.368, 0.596] \\
\bottomrule
\end{tabular}
\end{table}

\begin{table}[h]
\centering
\caption{Trained probe: out-of-fold AUC of an L2 logistic regression on per-trace summaries of each privileged context's signal, DeepMath L5-7 / L7-10, both teachers. The reference block never beats its content-free placebos.}
\label{tab:pred-probe}
\scriptsize
\begin{tabular}{lcc}
\toprule
context & 1.7B teacher & 8B teacher \\
\midrule
length only & .54 / .56 & .54 / .56 \\
reference (dense) & .57 / .58 & .65 / .64 \\
permuted reference & .60 / .60 & .66 / .71 \\
irrelevant reference (placebo) & .57 / .58 & .70 / .58 \\
answer only (sparse) & .69 / .75 & .78 / .85 \\
critique & .71 / .66 & .68 / .69 \\
placebo critique & .63 / .56 & .61 / .54 \\
\bottomrule
\end{tabular}
\end{table}

\section{Critique Ablations: Full Results}
\label{app:ablation}
The ablation result of \S\ref{sec:pointing} is reported here with confidence
intervals (Table~\ref{tab:ablation-full}), as paired per-rollout drops in which
each rollout's complete critique is compared with its own ablated critique
(Table~\ref{tab:ablation-paired}), and with the same paired test on the
Olmo-3-7B-Think rollouts. The pattern is the same in every form: removing the
quotes lowers localization, removing the reasons lowers correction, and the
7B Olmo-3 student reproduces it.

\begin{table}[h]
\centering
\caption{Ablating the critique, with 95\% confidence intervals: error-span Z when the gpt-5.2 critique in the teacher's context is complete (quote + reason), stripped of its quotes (reason only), or stripped of its reasons (quote only). Qwen3-1.7B rollouts, scale-up sample; scoring teacher Qwen3-1.7B in the upper block and Qwen3-8B in the lower. Critique minus donor-critique placebo throughout.}
\label{tab:ablation-full}
\scriptsize
\setlength{\tabcolsep}{3pt}
\begin{tabular}{llrlrl}
\toprule
critique & task & loc & 95\% CI & corr & 95\% CI \\
\midrule
quote + reason, 1.7B teacher & K\&K & $+0.75$ & $[+0.54,+0.97]$ & $-0.62$ & $[-0.86,-0.40]$ \\
 & L5-7 & $+0.99$ & $[+0.82,+1.16]$ & $-0.90$ & $[-1.14,-0.65]$ \\
 & L7-10 & $+0.78$ & $[+0.63,+0.92]$ & $-0.54$ & $[-0.71,-0.35]$ \\
reason only, 1.7B teacher & K\&K & $+0.48$ & $[+0.31,+0.67]$ & $-0.47$ & $[-0.67,-0.26]$ \\
 & L5-7 & $+0.75$ & $[+0.60,+0.89]$ & $-0.58$ & $[-0.81,-0.36]$ \\
 & L7-10 & $+0.61$ & $[+0.48,+0.75]$ & $-0.30$ & $[-0.45,-0.16]$ \\
quote only, 1.7B teacher & K\&K & $+0.71$ & $[+0.52,+0.90]$ & $-0.16$ & $[-0.39,+0.06]$ \\
 & L5-7 & $+0.94$ & $[+0.75,+1.14]$ & $-0.39$ & $[-0.58,-0.20]$ \\
 & L7-10 & $+0.69$ & $[+0.55,+0.85]$ & $-0.27$ & $[-0.43,-0.14]$ \\
\midrule
quote + reason, 8B teacher & K\&K & $+1.20$ & $[+0.95,+1.45]$ & $-0.79$ & $[-1.10,-0.49]$ \\
 & L5-7 & $+1.09$ & $[+0.91,+1.27]$ & $-0.96$ & $[-1.25,-0.68]$ \\
 & L7-10 & $+0.91$ & $[+0.77,+1.05]$ & $-0.62$ & $[-0.79,-0.45]$ \\
quote only, 8B teacher & K\&K & $+0.82$ & $[+0.63,+1.02]$ & $+0.21$ & $[-0.03,+0.45]$ \\
 & L5-7 & $+1.17$ & $[+0.98,+1.36]$ & $-0.22$ & $[-0.45,-0.01]$ \\
 & L7-10 & $+0.90$ & $[+0.75,+1.05]$ & $+0.03$ & $[-0.10,+0.17]$ \\
\bottomrule
\end{tabular}
\end{table}

\begin{table}[h]
\centering
\caption{Paired per-rollout ablation drops with 95\% CI: error-span Z of the complete critique minus that of the ablated critique on the same rollouts, gpt-5.2 critiques. Upper block: Qwen3-1.7B rollouts and teacher, scale-up sample. Lower block: Olmo-3-7B-Think rollouts and teacher, big set. A positive drop on localization or a negative drop on correction means the ablation removed signal; $^*$: CI excludes 0.}
\label{tab:ablation-paired}
\scriptsize
\setlength{\tabcolsep}{3pt}
\begin{tabular}{lllll}
\toprule
student & signal & ablation & K\&K & L5-7 / L7-10 \\
\midrule
Qwen3-1.7B & localization & remove quote & $+0.30^*$ & $+0.35^*$ / $+0.19^*$ \\
 & & remove reason & $-0.00$ & $+0.09$ / $+0.02$ \\
 & correction & remove quote & $-0.14$ & $-0.39^*$ / $-0.27^*$ \\
 & & remove reason & $-0.35^*$ & $-0.44^*$ / $-0.32^*$ \\
\midrule
Olmo-3-7B-Think & localization & remove quote & $+0.28^*$ $[+0.15,+0.42]$ & $+0.27^*$ $[+0.19,+0.34]$ / $+0.22^*$ $[+0.16,+0.28]$ \\
 & & remove reason & $+0.09$ $[-0.06,+0.25]$ & $+0.05$ $[-0.04,+0.14]$ / $+0.01$ \\
 & correction & remove quote & $-0.05$ & $-0.22^*$ $[-0.32,-0.12]$ / $-0.15^*$ $[-0.23,-0.08]$ \\
 & & remove reason & $-0.13$ & $-0.13^*$ $[-0.25,-0.02]$ / $-0.20^*$ \\
\bottomrule
\end{tabular}
\end{table}

\section{Replication on Olmo-3-7B-Think and Olmo-3-32B-Think}
\label{app:olmo}
The error-span measurements of the main text are repeated on a second model
family, with the student, the scoring teacher, and the self-critic all the same
model: Olmo-3-7B-Think on a set comparable in size to the Qwen scale-up, and
Olmo-3-32B-Think on a smaller set. Table~\ref{tab:olmo-setup} lists what
differs from the Qwen pipeline. The Olmo-3 models reason at greater length, so
their rollouts are sampled with a 32{,}768-token budget; the gold-span grader is
claude-sonnet-5 rather than a DeepSeek model, which leaves gpt-5.2 as the only
jury independent of the grader; part of the K\&K references were derived by
gpt-5.2 and verified against the ground truth, and the 32B study draws its
K\&K puzzles from the 632 that have a verified derivation. Everything else,
the prompts, the placebo constructions, the probe, and the error-span Z
readout, is shared.

The result replicates (Table~\ref{tab:olmo-spanz}): on both Olmo-3 models the
reference-conditioned teacher does not react at the errors, and the
critique-conditioned teacher localizes and corrects on every task. The 32B
model is the better self-critic: its own critiques localize on all three tasks
and correct on DeepMath 7-10, where the 7B's localize on the two math tasks
only. Table~\ref{tab:gate-full} gives the critique localization gate for both
models next to the full Qwen sweep; the Olmo-3 rollouts are two to three times
longer in sentences, so their absolute rates are comparable only within a
model.

\begin{table}[h]
\centering
\caption{The three error-span studies side by side: models, rollout sampling, gold-span pipeline, reference solutions, and the number of problems and rollouts that enter error-span Z on each task (Knights \& Knaves / DeepMath 5-7 / DeepMath 7-10).}
\label{tab:olmo-setup}
\scriptsize
\setlength{\tabcolsep}{3pt}
\resizebox{\linewidth}{!}{%
\begin{tabular}{llll}
\toprule
 & Qwen3-1.7B (scale-up) & Olmo-3-7B-Think (big set) & Olmo-3-32B-Think (minimal set) \\
\midrule
student = teacher = self-critic & yes (plus 8B teacher / critic) & yes & yes \\
rollout cap & 16{,}384 & 32{,}768 & 32{,}768 \\
sampling & temp 1.0, top-p 0.95, top-k 20, $n{=}8$ & same & same \\
gold-span grader & deepseek-reasoner / v4-flash & claude-sonnet-5 & claude-sonnet-5 \\
juries & gpt-5.2 + claude-sonnet-5 & same (gpt-5.2 independent) & same \\
oracle critiques & gpt-5.2, same prompt & gpt-5.2, same prompt & gpt-5.2, same prompt \\
K\&K references & verified deepseek derivations & 452 deepseek + 180 gpt-5.2, verified & 632 puzzles with a verified derivation \\
problems in error-span Z (K\&K / L5-7 / L7-10) & 355 / 516 / 794 & 197 / 398 / 631 & 114 / 111 / 133 \\
rollouts in error-span Z & 317 / 448 / 711 & 271 / 584 / 958 & 160 / 152 / 208 \\
\bottomrule
\end{tabular}}
\end{table}

\begin{table}[h]
\centering
\caption{Error-span Z at the first gold error with 95\% confidence intervals for the two Olmo-3 students, each scoring its own rollouts: Olmo-3-7B-Think on the big set and Olmo-3-32B-Think on the minimal set. Rows: reference solution in context; gpt-5.2 critique in context; the student's own critique in context. Localization on log content KL, correction on $\dlogp$ of the student's token; $n$ = rollouts entering the cell.}
\label{tab:olmo-spanz}
\scriptsize
\setlength{\tabcolsep}{3pt}
\begin{tabular}{lllrlrlr}
\toprule
student & conditioning & task & loc & 95\% CI & corr & 95\% CI & $n$ \\
\midrule
Olmo-3-7B & reference & K\&K & $+0.20$ & $[+0.01,+0.39]$ & $-0.00$ & $[-0.18,+0.18]$ & 271 \\
 & & L5-7 & $+0.11$ & $[-0.03,+0.25]$ & $-0.08$ & $[-0.23,+0.08]$ & 584 \\
 & & L7-10 & $-0.01$ & $[-0.11,+0.09]$ & $-0.03$ & $[-0.14,+0.08]$ & 958 \\
 & critique, oracle & K\&K & $+0.63$ & $[+0.39,+0.87]$ & $-0.29$ & $[-0.55,-0.03]$ & 271 \\
 & & L5-7 & $+0.89$ & $[+0.72,+1.07]$ & $-0.59$ & $[-0.78,-0.40]$ & 584 \\
 & & L7-10 & $+0.88$ & $[+0.75,+1.01]$ & $-0.49$ & $[-0.63,-0.33]$ & 958 \\
 & critique, self & K\&K & $+0.13$ & $[-0.08,+0.34]$ & $-0.14$ & $[-0.32,+0.04]$ & 271 \\
 & & L5-7 & $+0.41$ & $[+0.27,+0.56]$ & $-0.18$ & $[-0.32,-0.04]$ & 581 \\
 & & L7-10 & $+0.33$ & $[+0.21,+0.44]$ & $-0.14$ & $[-0.25,-0.03]$ & 953 \\
\midrule
Olmo-3-32B & reference & K\&K & $+0.09$ & $[-0.17,+0.35]$ & $-0.41$ & $[-0.72,-0.10]$ & 160 \\
 & & L5-7 & $+0.21$ & $[-0.06,+0.49]$ & $-0.09$ & $[-0.33,+0.14]$ & 152 \\
 & & L7-10 & $+0.15$ & $[-0.09,+0.37]$ & $-0.06$ & $[-0.32,+0.19]$ & 208 \\
 & critique, oracle & K\&K & $+0.76$ & $[+0.46,+1.07]$ & $-0.43$ & $[-0.79,-0.07]$ & 160 \\
 & & L5-7 & $+0.96$ & $[+0.66,+1.27]$ & $-0.50$ & $[-0.83,-0.19]$ & 152 \\
 & & L7-10 & $+1.20$ & $[+0.89,+1.48]$ & $-0.62$ & $[-0.99,-0.26]$ & 208 \\
 & critique, self & K\&K & $+0.47$ & $[+0.21,+0.72]$ & $-0.01$ & $[-0.30,+0.25]$ & 160 \\
 & & L5-7 & $+0.54$ & $[+0.26,+0.83]$ & $-0.39$ & $[-0.85,+0.05]$ & 150 \\
 & & L7-10 & $+0.40$ & $[+0.13,+0.65]$ & $-0.42$ & $[-0.73,-0.12]$ & 206 \\
\bottomrule
\end{tabular}
\end{table}

\begin{table}[h]
\centering
\caption{Critique localization gate for the two Olmo-3 students (oracle and self-critic) and the full window sweep for the Qwen3-1.7B rollouts. Percent of wrong rollouts with a gold span for which a quote lands within $w$ sentences of a gold error sentence; lenient tier; relaxed parse for self-critics. Olmo-3 rollouts run 800--1{,}500 sentences against 450--550 for Qwen, so absolute rates are comparable only within a student.}
\label{app:gate}
\label{tab:gate-full}
\scriptsize
\setlength{\tabcolsep}{3pt}
\resizebox{\linewidth}{!}{%
\begin{tabular}{llccccccc}
\toprule
student & critic & task & $\pm2$ & $\pm5$ & $\pm10$ & med.\ dist & quote-match & parse \\
\midrule
Qwen3-1.7B & gpt-5.2 (oracle) & K\&K / L5-7 / L7-10 & 48 / 54 / 59 & 57 / 62 / 67 & 65 / 69 / 73 & 3 / 2 / 1 & 98 / 96 / 95 & 100 \\
 & Qwen3-8B & & 15 / 13 / 16 & 26 / 22 / 27 & 40 / 31 / 37 & 17 / 18 / 13 & 96 / 82 / 83 & 99 / 86 / 85 \\
 & Qwen3-1.7B self, relaxed & & 9 / 12 / 13 & 20 / 18 / 21 & 33 / 26 / 31 & 16 / 23 / 16 & 77 / 64 / 68 & 90 / 83 / 83 \\
 & Qwen3-1.7B self, strict & & 4 / 8 / 9 & 9 / 13 / 15 & 16 / 19 / 22 & 12 / 16 / 13 & 33 / 42 / 45 & 43 / 54 / 53 \\
\midrule
Olmo-3-7B & gpt-5.2 (oracle) & & 31 / 37 / 46 & 39 / 45 / 55 & 43 / 51 / 64 & 16 / 8 / 3 & 97 / 89 / 94 & 100 \\
 & Olmo-3-7B self, relaxed & & 3 / 6 / 7 & 7 / 9 / 12 & 10 / 12 / 17 & 71 / 61 / 42 & 74 / 58 / 61 & 92 / 84 / 87 \\
\midrule
Olmo-3-32B & gpt-5.2 (oracle) & & 26 / 32 / 47 & 33 / 38 / 56 & 42 / 46 / 62 & 22 / 13 / 3 & 97 / 86 / 95 & 100 \\
 & Olmo-3-32B self, relaxed & & 4 / 9 / 13 & 8 / 11 / 21 & 12 / 17 / 24 & 110 / 76 / 51 & 87 / 80 / 87 & 98 / 93 / 97 \\
\bottomrule
\end{tabular}}
\end{table}

\section{Validation on Human-Annotated Error Spans}
\label{app:human}

Our main analysis uses error spans labelled by an LLM jury. To validate the result on human annotations, we repeat the analysis on ProcessBench \citep{zheng2025processbenchidentifyingprocesserrors}, which provides human-labelled first-error steps for solutions generated by Qwen2.5-7B-Instruct and Qwen2.5-Math-7B-Instruct. We score each solution with the model that generated it and compare reference conditioning against an oracle critique from gpt-5 \citep{openai2025gpt5systemcard}, generated without access to the human label. Table~\ref{tab:human} reproduces the main pattern of Table~\ref{tab:spanz}: the reference is neither localized nor corrective, while critique conditioning localizes the error for both models. Correction is clear for Qwen2.5-7B-Instruct and initially weak for Qwen2.5-Math-7B-Instruct; consistent with \S\ref{sec:pointing}, removing tokens quoted by the critique makes the signal corrective for both. More generally, correction appears when the critique points inside the human-labelled error span and disappears when it points elsewhere. Thus, human annotations reproduce the same picture: reference-conditioned scoring remains flat, while critique-conditioned scoring reacts where the critique identifies the error.

\begin{table}[t]
\centering
\caption{\textbf{On human-labelled error steps, the reference is flat and the
critique reacts.} ProcessBench, each solution teacher-forced through the
Qwen2.5 model that wrote it. Error-step Z as in Table~\ref{tab:spanz}: loc =
log content KL (positive = reacts at the labelled step), corr = $\dlogp$ of
the student's token (negative = corrective), hint minus placebo. Bold = 95\%
CI excludes 0. The critique is written by gpt-5 without the label.}
\label{tab:human}
\small
\setlength{\tabcolsep}{5pt}
\begin{tabular}{lrrrr}
\toprule
 & \multicolumn{2}{c}{Qwen2.5-7B-Instruct} & \multicolumn{2}{c}{Qwen2.5-Math-7B} \\
\cmidrule(lr){2-3}\cmidrule(lr){4-5}
conditioning & loc & corr & loc & corr \\
\midrule
reference & $+0.18$ & $-0.34$ & $+0.27$ & $-0.26$ \\
critique, oracle & $\mathbf{+1.25}$ & $\mathbf{-0.55}$ & $\mathbf{+0.80}$ & $+0.13$ \\
\bottomrule
\end{tabular}
\end{table}

\FloatBarrier

\end{document}